# ECA-RuleML: An Approach combining ECA Rules with temporal interval-based KR Event/Action Logics and Transactional Update Logics

Adrian Paschke

[1]Internet-based Information Systems, Technische Universität München
Adrian.Paschke@gmx.de

## Abstract

An important problem to be addressed within Event-Driven Architecture (EDA) is how to correctly and efficiently capture and process the event/action-based logic. This paper endeavors to bridge the gap between the Knowledge Representation (KR) approaches based on durable events/actions and such formalisms as event calculus, on one hand, and event-condition-action (ECA) reaction rules extending the approach of active databases that view events as instantaneous occurrences and/or sequences of events, on the other. We propose formalism based on reaction rules (ECA rules) and a novel interval-based event logic and present concrete RuleML-based syntax, semantics and implementation. We further evaluate this approach theoretically, experimentally and on an example derived from common industry use cases and illustrate its benefits.

## 1. Motivation

Event-driven applications based on reactive rules and in particular ECA rules which trigger actions as a response to the detection of events have been extensively studied during the 1990s. Stemming from the early days of programming language where system events were used for interrupt and exception handling, active event-driven rules have received great attention in different areas such as active databases [1, 2] which started in the late 1980s, real-time applications and system and network management tools which emerged in the early 1990s as well as publish-subscribe systems [3] which appeared in the late 1990s. Recently, there has been an increased interest in industry and academia in event-driven mechanisms and high-level Event-Driven Architectures (EDA). (Pro-)active real-time or just-in-time reactions to events are a key factor in upcoming agile and flexible IT infrastructures, distributed loosely coupled service oriented environments or new business models such as On-Demand or Utility computing. Industry trends such as Real-Time Enterprise (RTE), Business Activity Management (BAM) or Business Performance Management and closely related areas such as Service Level Management (SLM) with monitoring and enforcing Service Level Agreements (SLAs) are business drivers for this renewed interest. Another strong demand for event processing functionalities comes from the web community, in particular in the area of Semantic Web and Rule Markup Languages (e.g. RuleML[1]).

Active databases in their attempt to combine techniques from expert systems and databases to support automatic triggering of rules in response to events and to monitor state changes in database systems have intensively explored and developed the **ECA paradigm** and **event algebras** to compute complex events. In a nutshell, this paradigm states that an ECA rule autonomously reacts to actively or passively detected simple or complex events by evaluating a condition or a set of conditions and by executing a reaction whenever the event happens and the condition(s) is true: "*On Event if Condition do Action*".

A different approach to events and actions – the so called **KR event/action logics** - which has for the most part proceeded separately has the origin in the area of artificial intelligence (AI), knowledge representation (KR) and logic programming (LP). Here the focus is on the development of axioms to formalize the notions of actions/events and causality, where events/actions are characterized in terms of necessary and sufficient conditions for their occurrences. Instead of detecting the events as they occur in active databases at a single point in time, the KR approach to events focuses on the inferences that can be made from the fact that certain events are known to have occurred or are planned to happen in future. This leads to different views and terminologies on event/actions definition and event/action processing in these domains. In typical event algebras of active database systems such as Snoop [4], events are defined in terms of their detection time conditions and consumed to build up complex events, i.e., events are viewed as occurring at their time of detection and hence as transient and instantaneous, while in the KR approaches such as Event Calculus, events are treated as non-transient, durative facts which initiate or terminate validity intervals, i.e. have an effect which holds at an interval. Solely treating (complex) events in terms of transient atomic instants occurring at a single point in time, rather than durable occurrences which hold over an interval, leads to some logical problems and unintended semantics for several event algebra operators. The non-transient events are simply detected and consumed, whereas the information about the effects they initiate/terminate and how far into the past (or future) that effects extend is lost.

Pure event-driven architectures, active database systems, production rule systems or message-driven event notification systems (ENS) often only implement procedural semantics but lack a precise formal semantics. This is a serious omission for many real-world applications where validation and traceability of the effects of events and the triggered actions are crucial, e.g., reactive rules are often being used as a programming language to describe real-world active decision logic (e.g., business rules, contract rules, policies) and create production systems upon, which demands verification and traceability support. For this reason, it is important to provide also a sound and complete formal semantics that can help determining the reliability of the results produced by the event processing logic. KR Event/Action logics and their implementations as (meta) logic programs have a great potential to fill this gap between active procedural event processing stemming from active databases and KR-based event/action logics. In this paper, we present a framework for combining both approaches and demonstrate the benefits of this combination. We further show how this approach fits into RuleML, a standardization initiative for describing different rule types on the (Semantic) Web and develop the ECA Rule Markup Language (ECA-RuleML). The further paper is structured as follows: In section 2 we define basic concepts in active event processing relevant to our work. We then show in section 3 that typical event algebras in the active database domain treat complex events as single occurrences at a particular point in time, which leads to unintended semantics for several of their operators. Based on this observation, we evolve an interval-based variant of the Event Calculus which treats complex events as a sequence of intervals of all its component events. We redefine typical algebra operators of active database systems with this new interval-based logic formalism and show that the detected problems and unintended semantics do not occur here. In section 4, we describe an approach to event processing based on active ECA rules in Logic Programming and relate it to the KR event/action logics approach evolved in section 3. In section 5, we present the ECA Rule Markup Language (ECA-RuleML) that enables serialization of reactive rules (ECA rules) and event algebra operators in RuleML. In section 6, we evaluate our approach theoretically by means of asymptotic worst case analysis and experimentally with regards to computational complexity and demonstrate its adequacy with an example derived from a common industrial use case. In section 7, we discuss related work and conclude our work in section 8 with a summary and an outlook on future steps.

[1] http://www.ruleml.org

## 2. Basic Concepts and History in Active Event / Action Processing

Before we discuss the details of the implementation of the LP based ECA processor, the event algebra based on a KR event/action logics formalism, we first give a review of the history of event / action processing in different domains and then outline some relevant concepts and definitions of our active event/action processing approach.

### 2.1 Overview on Event / Action Processing

#### Active Databases and ECA Rule Systems

Active databases are an important research topic due to the fact that they find many applications in real world systems and many commercial databases systems have been extended to allow the user to express **active rules** whose execution can update the database and trigger the further execution of active rules leading to a cascading sequence of updates (often modelled in terms of execution programs). Several active database systems have been developed, e.g. ACCOOD [5], Chimera [6], ADL [7], COMPOSE [8], NAOS [9], HiPac [10]. These systems mainly treat event detection and processing purely procedural and often focus on specific aspects. In this spirit of procedural ECA formalisms are also systems such as AMIT [11], RuleCore [12] or JEDI [13]. Several papers discuss formal aspects of active databases on a more general level – see e.g. [14] for an overview. Several event algebras have been developed, e.g. Snoop [4], SAMOS [15], ODE [8]: The object database ODE [8] implements event-detection mechanism using finite state automata. SAMOS [16] combines active and object-oriented features in a single framework using colored Petri nets. Associated with primitive event types are a number of parameter-value pairs by which events of that kind are detected. SAMOS does not allow simultaneous atomic events. Snoop [4] is an event specification language which defines different restriction policies that can be applied to the operators of the algebra. Complex events are strictly ordered and cannot occur simultaneously. The detection mechanism is based on trees corresponding to the event expressions, where primitive event occurrences are inserted at the leaves and propagated upwards in the trees as they cause more complex events to occur. There has been a lot of research and development concerning knowledge updates in active rules (execution models) in the area of active databases and several techniques based on syntactic (e.g. triggering graphs [17] or activation graphs [18]) and semantics analysis (e.g. [19], [20]) of rules have been proposed to ensure termination of active rules (no cycles between rules) and confluence of update programs (always one unique minimal outcome). The combination of deductive and active rules has been also investigated in different approaches manly based on the simulation of active rules by means of deductive rules [21-23].

#### Production Rule Systems and Update Rule Programs

The treatment of active rules in active databases is to some extend similar to the **forward-chaining production rules system paradigm** in artificial intelligence (AI) research [24]. In fact, triggers, active rules and integrity constraints, which are common in active DBMS, are often implemented in a similar fashion to forward-chaining production rules where the changes in the conditions due to update actions such as "*assert*" or "*retract*" on the internal database are considered as implicit events leading to further update actions and to a sequence of "firing" production rules, i.e.: "*if Condition then Action*". There are many forward-chaining implementations in the area of deductive databases and many well-known forward-reasoning engines for production rules such as ILOG's commercial jRules system, Fair Isaac/Blaze Advisor, CA Aion, Haley, ESI Logist or popular open source solutions such as OPS5, CLIPS or Jess which are based on the RETE algorithm. In a nutshell, this algorithm keeps the derivation structure in memory and propagates changes in the fact and rule base. This algorithm can be very effective, e.g. if you just want to find out what new facts are true or when you have a small set of initial facts and when there tend to be lots of different rules which allow you to draw the same conclusion. This might be also on reason why production rules have become very popular as a widely used technique to implement large expert systems in the 1980s for diverse domains such as troubleshooting in telecommunication networks or computer configuration systems. Classical production rule systems and most database implementations of production rules [25-27] typically have an operational or execution semantics defined for them, but lack a precise theoretical foundation and do not have a formal semantics. Although production rules might simulate derivation rules via asserting a conclusion as consequence of the proved condition, i.e. "*if Condition then assert Conclusion*", the classical production rule languages such as OPS5 are less expressive since they lack a clear declarative semantics, suffer from termination and confluence problems of their execution sequences and typically do not support expressive non-monotonic features such as classical or negation-as-finite failure or preferences, which makes it sometimes hard to express certain real life problems in a natural and simple way. However, several extensions to this core production systems paradigm have been made which introduce e.g. negations (classical and negation-as-finite failure) [28] and provide declarative semantics for certain subclasses of production rules systems such as stratified production rules. It has been shown that such stratified production systems have a declarative semantics defined via their corresponding logic program (LP) into which they can be transformed [29] and that the well-founded, stable or preferred semantics for production rule systems coincide in the class of stratified production systems [28]. Stratification can be implemented on top of classical production rules in from of priority assignments between rules or by means of transformations into the corresponding classical ones. The strict definition of stratification for production rule systems has been further relaxed in [30] which defines an execution semantics for update rule programs based upon a monotonic fixpoint operator and a declarative semantics via transformation of the update program into a normal LP with stable model semantics. Closely related are also logical update languages such as transaction logics and in particular serial Horn programs, where the serial Horn rule body is a sequential execution of actions in combination with standard Horn pre-/post conditions. [31] These serial rules can be processed top-down or bottom-up and hence are closely related to the production rules style of "*condition* → *update action*". Several approaches in the active database domain also draw on transformations of active rules into LP derivation rules, in order to exploit the formal declarative semantics of LPs to overcome confluence and termination problems of active rule execution sequences. [32-34]. The combination of deductive and active rules has been also investigated in different approaches mainly based on the simulation of active rules by means of deductive rules. [21, 22] Moreover, there are approaches which directly build reactive rules on top of LP derivation rules such as the Event-Condition-Action Logic Programming language (ECA-LP) which enables a homogeneous representation of ECA rules and derivation rules [35].

**Event Notification Systems and Reaction Rule Interchange Languages**

Recently, a strong demand for event/action processing functionalities comes from the web community, in particular in the area of Semantic Web and Rule Markup and the upcoming W3c Rule Interchange Language (e.g. RuleML[2] or RIF[3]). In distributed environments such as the (Semantic) Web with independent agent / system nodes and open or closed domain boundaries event processing is often done using **event notification and communication mechanisms**. Systems either communicate events in terms of messages according to a predefined or negotiated communication/coordination protocol [36] and possibly using a particular language such as the FIPA Agent Communication Language (FIPA-ACL) or they subscribe to publishing event notification servers which actively distribute events (push) to the subscribed and listening agents using e.g. common format such as the common base event format [37]. Typically the interest here is in the particular event sequence which possibly follows a certain communication or coordination protocol, rather than in single event occurrences which trigger immediate reactions as in the active database trigger or ECA rules.

**Temporal KR Event / Action / Transition Logic Systems**

A fourth dimension to events and actions which has for the most part proceeded separately has the origin in the area of knowledge representation (KR) and logic programming (LP) with close relations to the formalisms of process and transition logics. Here the focus is on the development of axioms to formalize the notions of actions resp. events and causality, where events are characterized in terms of necessary and suf-ficient conditions for their occurrences and where events/actions have an effect on the actual knowledge states, i.e. they transit states into other states and initiate / terminate changeable properties called **fluents**. Instead of detecting the events as they occur as in the active database domain, the KR approach to events/actions focuses on the inferences that can be made from the fact that certain events are known to have occurred or are planned to happen in future. This has led to different views and terminologies on event/action definition and event processing in the **temporal event/action logics** domain. Reasoning about events, actions and change is a fundamental area of research in AI since the events/actions are pervasive aspects of the world in which agents operate enabling retrospective reasoning but also prespective planning. A huge number of formalisms for deductive but also abductive reasoning about events, actions and change have been developed. The common denominator to all this formalisms and systems is the notion of states a.k.a. fluents [38] which are changed or transit due to occurred or planned events/actions. Among them are the event calculus [39] and variants such as the interval-based Event Calculus (see section 3.3), the situation calculus [40, 41], features and fluents [38], various (temporal) action languages [42-46], fluent calculi [47, 48] and versatile event logics [49]. Most of these formalisms have been developed in relative isolation and the relationships between them have only been partially studied, e.g. between situation calculus and event calculus or temporal action logics (TAL) which has its origins in the features and fluents framework and the event calculus. Closely related and also based on the notion of (complex) events, actions and states with abstract models for state transitions and parallel execution processes are various process algebras like TCC [50], CSS [51] or CSP [52], (labelled) transition logics (LTL) and (action) computation tree logics (ACTL) [53, 54]. Related are also update languages [22, 35, 55-61] and transaction logics [31] which address updates of logic programs where the updates can be considered as actions which transit the initial program (knowledge state/base) to a new extended or reduced state hence leading to a sequence of evolved knowledge states. Many of these update languages also try to provide meaning to such dynamic logic programs (DLPs). However, unlike ECA languages and production rules these languages typically do not provide complex event / action processing features and exclude external calls with side effects via event notifications or procedural calls.

[2] http://www.ruleml.org

[3] W3C RIF

## 2.2. Basic Concepts in Event / Action Processing

**Definition 1:** *Atomic Events, Event Type Pattern Definitions and Event Context*

A raw event (a.k.a. atomic or primitive event) is defined as an instantaneous (occurs in a specific point in time), significant (relates to a context), atomic occurrence (it cannot be further dismantled and happens completely or not at all):

occurs(e,t).  event e occurs at time point t

We distinguish between event instances (simply called events) which occur and event type pattern definitions. An event type pattern definition (or simply event definition or event type) describes the structure of an (atomic or complex) event, i.e. it describes its detection condition(s). We use (derivation) rules to formalize these detection conditions:

detect(e,T) :- <detection conditions>,

which means that an event instance *e* is detected at time *T*, whenever the detection conditions defined in the rules' body are fulfilled. In our interval-based approach *T* is an occurrence interval *[T1,T2]* during which the complex event *e* occurs. An atomic event occurs at *[T,T]*. A concrete "instantiation" of a type pattern is a specific event instance, which is derived (detected) from the detection conditions defined within the rules body. This can be, e.g., a particular fact becoming true, an external event in an monitored system, a communication event within a conversation, a state transition from one state to another such as knowledge updates, transactions (e.g., begin, commit, abort, rollback), temporal events (e.g., at 6 p.m., every 10 minutes), etc.

Typically, events occur in a context that is often relevant to the execution of the other parts of the ECA rules, i.e., event processing is done within a context. A context can have different characteristics such as:

- Temporal characteristic designates information with a temporal perspective, e.g., service availability within one month or within 60 minutes from X.
- Spatial characteristic designates information with a location perspective, e.g. message reached end-point.
- State characteristic designates information with a state perspective, e.g., low average response time.
- Semantic characteristic designates information about a specific object or entity, e.g. persons that belong to the same role or messages that belongs to same conversation.

To capture the local context of an event, we use variables, i.e., the context information is bound to variables which can be reused in the subsequent parts of an ECA rule, e.g., in the action part.

**Definition 2** *Complex Events and Event Algebra*

A complex event type (i.e. the detection condition of a complex event) is built from occurred atomic or other complex event instances according to the operators of an event algebra. The included events are called components while the resulting complex event is the parent event. The first event instance contributing to the detection of a complex event is called initiator, where the last is the terminator; all others are called interiors. We define the operators of the event algebra in terms of LP event logics formalizations. For example, a straightforward definition of the sequence operator is that the sequence *(e1;e2)*, which defines the detection condition for the complex event *e3*, should be detected whenever *e1* occurs and then *e2* occurs, i.e. the occurrence time of *e1* is before the occurrence time of *e2*: *T1<T2*[4]:

detect(e3,T2) :- occurs(e1,T1), occurs(e2,T2), T1<T2.

An inference engine is used to derive (detect) all occurrences of *e3* from the recorded event instances of *e1* and *e2,* i.e. from the "occurs" facts within the knowledge base (KB). After this event instance selection the detected complex event(s) must be remembered in order to reuse it for further detection of other complex events and all component events which have contributed so far should be consumed, i.e. they are removed from the KB and can no longer contribute to the detection of other events.

**Definition 3** *Event Processing*
Event processing describes the process of selecting and composing complex events from raw events (event derivation), situation detection (detecting transitions in the universe that requires reaction either "reactive" or "proactive") and triggering actions as a consequence of the detected situation (complex event + conditional context). Examples are: "*If more than three outages occur then alert*" or "*If a department D is retracted from the database than retract all associated employees E*". Events can be processed in real time without persistence (short-term) or processed in retrospect as a computation of persistent earlier events (long-term), but also in aggregated from, i.e., new (raw) events are directly added to the aggregation which is persistent. According to the ECA paradigm event processing can be conditional, i.e. certain conditions must hold before an action is triggered. Event processing can be done either actively (pull-model), i.e. based on actively monitoring the environment and, upon the detection of an event, trigger a reaction, or passively (push-model), i.e., the event occurrences are detected by an external component and pushed to the event system for further processing.

Most event algebras in the active database domain treat (complex) events as instantaneous occurrences at a particular point in time, e.g. *(A;B)* in Snoop [4] is associated with the occurrence time of *B* (the terminator), i.e. the complex event is detected at the time point when *B* occurs. In the next section we illustrate that this results in unintended semantics for some compositions in these algebras. We further demonstrate that these problems do not arise, if we associate the detection of a complex event with the maximum validity interval (MVI) in which it occurs, i.e., the interval in which all component events occurred, rather then the time of detection given by the terminator timestamp. This interval-based treatment of event definitions is a widespread assumption in the KR community and typical event logics such as the Event Calculus [39] enable temporal reasoning to derive the effects of events which hold at an interval beginning with an "initiating" event and ending with a "terminating" event. We extend the Event Calculus formalization of temporal event based reasoning with the notion of event intervals which hold between time intervals *[t1,t2]*. We use this new interval-based EC variant to redefine typical event algebra operators and investigate how far it is possible to reconcile the active database approach based on events detectable at a single point in time with the durative, inference-based treatment of events, occurring within an interval, in the sense of KR event/action logics.

[4] In the next section we show that this initial formalization is to general and needs further refinement.

### 2.3. A Classification of the Event / Action / Processing Space

***I) Classification of the Event Space***

**1. Processing**
(situation detection or event/action computation / reasoning)

- **Short term**: Transient, non-persistent, real-time selection and consumption (e.g. triggers, ECA rules): ***immediate reaction***
- **Long term**: Transient, persistent events, typically processed in retrospective e.g. via KR event reasoning or event algebra computations on event sequence history; but also prospective planning / proactive, e.g. KR abductive planning: ***defered or retrospective/prospective***
- **Complex event processing**: computation of complex events from event sequence histories of previously detected raw or other computed complex event (event selection and possible consumption) or transitions (e.g. dynamic LPs or state machines); typically by means of event algebra operators (event definition) (e.g. ECA rules and active rules, i.e. sequences of rules which trigger other rules via knowledge/state updates leading to knowledge state transitions)
- **Deterministic vs. non-deterministic**: simultaneous occurred events give rise to only one model or two or more models
- **Active vs. Passive**: actively detect / compute / reason event (e.g. via monitoring, sensing akin to periodic pull model or on-demand retrieve queries) vs. passively listen / wait for incoming events or internal changes (akin to push models e.g. publish-subscribe):

**2. Type**

- **Flat vs. semi-structured compound data structure/type**, e.g. simple String representations or complex objects with or without attributes, functions and variables
- **Primitive vs. complex**, e.g. atomic, raw event or complex derived/computed event
- **Temporal**: Absolute (e.g. calendar dates, clock times), relative/delayed (e.g. 5 minutes after …), durable (occurs over

an interval), durable with continuous, gradual change (e.g. clocks, countdowns, flows)

- **State or Situation**: flow oriented event (e.g. "server started", "fire alarm stopped")
- **Spatio / Location**: durable with continuous, gradual change (approaching an object, e.g. 5 meters before wall, "bottle half empty" )
- **Knowledge Producing**: changes agents knowledge belief and not the state of the external world, e.g. look at the program → effect

**3. Source**

- **Implicit** (changing conditions according to self-updates) vs. **explicit** (**internal** or **external** occurred/computed/detected events) (e.g. production rules vs. ECA rules)
- **By request** (query on database/knowledge base or call to external system) vs. **by trigger** (e.g. incoming event message, publish-subscribe, agent protocol / coordination)
- **Internal database/KB update events** (e.g. add, remove, update, retrieve) or **external explicit events** (inbound event messages, events detected by external systems): **belief update and revision**
- **Generated/Produced (e.g. phenomenon, derived action effects) vs. occurred (detected or received event)**

***II) Classification of the Action Space***

**Similar dimensions as for events (see above)**

**1. Temporal KR event/action perspective**:
(e.g. Event, Situation, Fluent Calculus, TAL)

- Actions with effects on changeable properties / states, i.e. actions ~ events
- Focus: reasoning on effects of events/actions on knowledge states and properties

**2. KR transaction, update, transition and (state) processing perspective**:
(e.g. transaction logics, dynamic LPs, LP update logics, transition logics, process algebra formalism)

- Internal knowledge self-updates of extensional KB (facts / data) and intensional KB (rules)
- External actions on external systems via (procedural) calls, outbound messages, triggering/effecting
- Transactional updates possibly safeguarded by post-conditional integrity constraints / test case tests
- Complex actions (sequences of actions) modeled by action algebras (~event algebras), e.g. delayed reactions, sequences of bulk updates, concurrent actions
- Focus: declarative semantics for internal transactional knowledge self-update sequences (dynamic programs)

**3. Event Messaging / Notification System perspective**

- Event/action messages (inbound / outbound messages)
- Often: agent / automated web) service communication; sometimes with broker, distributed environment, language primitives (e.g. FIPA ACL) and protocols; event notification systems, publish / subscribe
- Focus: often follow some protocol (negotiation and coordination protocols such as contract net) or publish-subscribe mechanism

**4. Production rules perspective:**
(e.g. OPS5, Clips, Jess, JBoss Rules/Drools, Fair Isaac Blaze Advisor, ILog Rules, CA Aion, Haley, ESI Logist )

- Mostly forward-directed non-deterministic operational semantics for Condition-Action rules
- Primitive update actions (assert, retract); update actions (interpreted as implicit events) lead to changing conditions which trigger further actions, leading to sequences of triggering production rules
- But: approaches to integrate negation-as-failure and declarative semantics exist, e.g. for subclasses of production rules systems such as stratified production rules with priority assignments or transformation of the PR program into a normal LP
- Related to serial Horn Rule Programs

**5. Active Database perspective**:
(e.g. ACCOOD, Chimera, ADL, COMPOSE, NAOS, HiPac)

- ECA paradigm: "*on Event and Condition do Action*"; mostly operational semantics
- Instantaneous, transient events and actions detected according to their detection time
- Complex events: event algebra (e.g. Snoop, SAMOS, COMPOSE) and active rules (sequences of self-triggering ECA rules)

***III) Classification of the Event / Action / State Processing respectively Reasoning Space***

**1. Event/Action Definition Phase**

- Definition of event/action pattern by event algebra
- Based on declarative formalization or procedural implementation
- Defined over an atomic instant or an interval of time, events/actions, situation, transition etc.
- 

**2. Event/Action Selection Phase**

- Defines selection function to select one event from several occurred events (stored in an event instance sequence e.g. in memory, database/KB) of a particular type, e.g. "*first*", "*last*"
- Crucial for the outcome of a reaction rule, since the events may contain different (context) information, e.g. different message payloads or sensing information
- **KR view**: Derivation over event/action history of happened or future planned events/actions
-

**3. Event/Action Consumption / Execution Phase**

- Defines which events are consumed after the detection of a complex event
- An event may contribute to the detection of several complex events, if it is not consumed
- Distinction in event messaging between "multiple receive" and "single receive"
- Events which can no longer contribute, e.g. are outdated, should be removed
- **KR view**: events/actions are not consumed but persist in the fact base
-

**4. State / Transition Processing**

- Actions might have an internal effect i.e. change the knowledge state leading to state transition from (pre)-condition state to post-condition state.
- The effect might be hypothetical (e.g. a hypothetical state via a computation) or persistent (update of the knowledge base),
- Actions might have an external side effect

➔ Separation of this phases is crucial for the outcome of a reaction rule based system since typically event occur in a context and interchange context date to the condition or action (e.g. via variables, data fields).

➔ Declarative configuration and semantics of different selection and consumption policies is desirably

## 3. An Event Algebra based on the interval-based Event Calculus

### 3.1. Unintended Semantics of Event Algebra Operators in the Active Database Domain

Typical event algebras in the active database domain, such as Snoop, define the following operations or variants of them:

- Sequence operator (;): the specified event instances have to occur in the order determined by this operator
- Disjunction operator (˅): at least one of the specified instances has to occur
- Conjunction operator (˄): the specified event instances can occur in any order, where the detection time is the timestamp of the latest occurred event
- Simultaneous operator (=): the specified instances have to occur simultaneously
- Negation operator (¬): the specified instance(s) are not allowed to occur in a given interval
- Quantification (Any): the complex event occurs when n events of the specified type have occurred
- Aperiodic Event Operator (Ap): The aperiodic operator *A* allows one to express the occurrence of an event *E2* within the interval defined by two other events *E1* and *E3: Ap (E2, E1, E3)*
- Periodic Event Operator (Per): The periodic operator *Per* is an event type which occurs every *t* time-steps in between *E1* and *E2*

The detection time of a complex event is typically the occurrence time of its terminating event (according to the defined selection and consumption policy). This leads to some inconsistencies and irregularities in typical event algebras as we will illustrate now:

1. Consider the following event type *Ap (A, B,C)* defined by the aperiodic operator *Ap* in Snoop, i.e. the complex event is detected when *A* occurs between *B* and *C*. An event instance sequence (EIS) *{a c b b}* will trigger two events of this type, according to the type specific instance sequence *Ap (A,B,C)={{a,b},{a,b}},* even though the event instances of *B* occur outside the detection interval defined by *a* and *c*.
2. The sequence *B;(A;C)* in Snoop is detected if *A* occurs first, and then *B* followed by *C, i.e. EIS={a, b, c}*, because the complex event *(A;C)* is detected with associated detection time of the terminator *c* and accordingly the event *b* occurs before the detected complex event *(A;C)*. However, this is not the intended semantics: Only a *EIS={b, a, c}* should lead to the detection of the complex event *B;(A;C)* and the correct event type pattern for *EIS={a, b, c}* should be *A;(B;C).* In the semantics of Snoop, which uses the detection time of the terminating event as occurence time of a complex event, both complex event definition *B;(A;C)* and *A(B;C)* are equal.
3. Consider the event type *(A;B) [A, C]* in SAMOS, i.e. the event of type *A* sequentially followed by *B* in between *A* and *C* causes the complex event to be detected. For an EIS *{a a b b b c}* one may expect (under a strict interpretation) that it does detect one (or at least two) complex event, because the sequence *(a;b)* occurs once (or twice) in the interval determined by *a* and *c*. However, the event detection algorithm used in SAMOS detects three events of this type, because it combines all occurrences of *a* and *b* in between *[A, C]* regardless of the sequence operator and occurrence intervals of *[A,B]* in between *[A,C]*. Although this might be acceptable in some domains, under a strict interpretation it violates the semantics of the sequence operator and the usual event consumption, where events do not contribute to several event detections but are consumed after they have contributed to a complex event. Moreover, it uses all event *a* and *c* to span several intervals *[A,C]* which are not terminated in between, e.g. because *a* occurs again or *b* followed by *a* occurs (i.e. the sequence definition is violated). This leads to inconsistencies in complex event detection under a strict interpretation.
4. Given the event type *(A;B)* ACCOD (Active Object Oriented Database System) causes the recognition of six complex events from the EIS: *{a a a b b}*. Hence, their definition of the sequence operator is concurrent with our straightforward logical formalization: (E1;E2) ≡ occurs(E1,T1), occurs(E2,T2), T1<T2. While this may be acceptable, or even desirable, in some applications, strictly speaking, if the events unify with the detection interval that is already progressing, i.e. which is initiated by *a*, multiple repeated occurrences of *a* should be ignored. In the strict case (strict intepretation) that *A* should follow *B* and a new *a* occurs, while the monitoring interval *[a,b]* is still expecting *b* to complete, the new event should be ignored.

All these problems arise from the fact that the events, in the active database sense, are simply detected and treated as if they occur at an atomic instant, in contrast to durative complex events, in the KR event logics sense, which occur over an extended interval, where the information about how far into the past their occurrence interval reaches is derived. To solve these problems, we redefine the typical event algebra operators by associating the occurrences of complex events with their detection intervals, i.e., we formalize the typical event algebra operators in terms of logical event detection conditions using a deductive event logic formalism - the Event Calculus. We first describe the basic concepts of the classical Event Calculus and then extend it with the notion of event intervals which occur over a time interval leading to a novel interval based Event Calculus variant which better supports our KR based event algebra.

### 3.2. Basic Event Calculus

The Event Calculus [39] is a deductive KR formalism for temporal reasoning about the effects of events. Rather than the detection of events as they occur, the Event Calculus approach on events concentrates on what inferences can be made from the fact that certain events are known to have occurred (or are planned to happen in future, thus providing a link to AI planning). It allows specifying rules on events which initiate or terminate an interval and derive the (maximum) validity periods of these intervals. It defines a model of change in which *events* happen at *time-points* and *initiate* and/or *terminate time-intervals* over which some *properties* (time-varying **fluents**) of the world hold. Time-points are unique points in time at which events take place instantaneously. The basic idea is to state that fluents (intervals) are true at particular time-points if they have been initiated by an event at some earlier time-point and not terminated by another event in the meantime. Similarly, a fluent is false at a particular time-point, if it has been previously terminated and not initiated in the meantime, i.e. the EC embodies a notion of default persistence according to which fluents are assumed to persist until an event occurs which terminates them. This principle follows the *axiom of intertia* first proposed by McCarthy and Hayes which says: "*Things normally tend to stay the same*" [62]. A central feature of the EC is that it establishes or assumes a time structure which is independent of any event occurrences. The time structure is usually assumed or stated to be linear although the underlying ideas can equally be applied to other temporal notions, e.g. branching structures, i.e. event occurrences can be provided with different temporal qualifications. Variants range from normal structures with absolute times and total ordering to loose ones with only relative times and partial ordering. Therefore, a central feature of the Event Calculus, in comparison to other event-based temporal logics such as Situation Calculus, is that an explicit linear time-structure, which is independent of any events, is assumed. The EC axioms describe when events occur (transient view) / happen (non-transient view) within this time structure and which properties (fluents) are initiated and/or terminated by these events under various circumstances:

Predicates:

| | |
|---|---|
| *occurs(E,T)* | *event E occurs at time interval T:=[T1,T2]* |
| *happens(E,T)* | *event E happens at time point T* |
| *initiates(E,F,T)* | *event E initiates fluent F for all time>T* |
| *terminates(E,F,T)* | *event E terminates fluent F for all time>T* |

Auxiliary predicates:

| | |
|---|---|
| *clipped(T1,F,T2)* | *fluent F is terminated between T1 and T2* |
| *declipped(T1,F,T2)* | *fluent F is initiated between T1 and T2* |

Domain independent axioms:

| | |
|---|---|
| *holdsAt(F,T)* | *fluent F holds at time point T* |
| *not(holdsAt(F,T1))* | *fluent F does not hold at time T1* |

Appendix D shows our optimized LP formulation of the basic EC holdsAt axiom. Note that the implementation of the EC axioms is fully declarative and is **completely un-typed**, i.e. we make no restrictions on the type of terms used within the EC axioms, i.e. an event might be a simple logical term, e.g. "*e1*", a complex term with variables "*e1(X,Y)*" or even a typed external procedural object, e.g. a Java object. "*rbsla.eca.Event()*", since we are using a Java resp. Description Logic typed logic (http://ibis.in.tum.de/staff/paschke/docs/ContractLog_OWL_12_05.pdf).

We distinguish between occurred events (*occurs(e,t)*) and happened events (*happens(e,t)*) to capture the semantics of volatile event instances consumed within active event processing (occurs) and non-transient KR events (happens) used for inferring traceable, retrospective results in order to build complex and reliable decision logic upon. We now extend the basic EC formulation with event intervals which occur between time intervals, leading to a novel interval based Event Calculus variant.

### 3.3. Interval-based Event Calculus

Complex events are typically durative, i.e. take time, and span a time interval between the initiator and the terminator of the complex event. In terms of the EC this means, that the occurrence of an (atomic) initiator event $e_i$, i.e., occurs($e_i$, [$t_{ei}$, $t_{ei}$]), initiates and the occurrence of a terminator event $e_t$, i.e., oc-

curs($e_t$, [$t_{et}$,$t_{et}$]), terminates the occurrence interval $[t_{ei},t_{et}]$ of the event interval $[e_i,e_t]$, where $t_{ei}<t_{et}$: initates($e_i$, [$e_i$,$e_t$], T). terminates($e_t$, [$e_i$,$e_t$], T).

Since we already know from the definition of the event interval which event is the initiator and which is the terminator the explicit initiates and terminates axioms can be omitted. However, there might be other events, the so called terminators, which are not allowed to occur between the event interval $[e_i,e_t]$, i.e. they break the interval. For example in a sequence definition *(a;b;c)* the event *c* will break the event interval *[a,b]*. These terminators might be explicitly and globally captured by the terminates axioms, e.g.: terminates(c,[a,b],T). We now reformulate the basic EC formalization to an interval-based EC where time intervals are the basic temporal concept. Here all events are regarded to occur in an time interval, i.e. an event interval *[e1,e2]* occurs during the time interval *[t1,t2]* where *t1* is the occurrence time of *e1* and *t2* is the occurrence time of *e2*. Note, that an atomic event occurs in the interval *[t,t]*, where *t* is the occurrence time of the atomic event. The basic holdsAt axiom used in the EC for temporal reasoning about fluents is redefined to the axiom *holdsInterval([E1,E2],[T1,T2])* to capture the semantics of event intervals which hold between a time interval – see also appendix E:

```
% free holdsInterval queries: holdsInterval(+, -) +=input ; - =output
% returns the time interval [T1,T2] for the event interval [E1,E2]
holdsInterval([E1,E2],[T11,T22]):-
        free(T11), free(T22),
        event([E1],[T11,T12]),
        event([E2],[T21,T22]),
        [T11,T12]<=[T21,T22],
        not(broken(T12,[E1,E2],T21).
% bound holdsInterval queries: holdsInterval(+, +) +=input ; - =output
% Test whether the event interval [E1,E2] holds between [T1,T2]
holdsInterval([E1,E2],[T1,T2]):-
        bound(T1), bound(T2),
        event([E1],[T11,T12]),
        T11<=T1,
        event([E2],[T21,T22]),
        T2<=T22
        not(broken(T1,[E1,E2],T2).
% atomic event interval with one event
holdsInterval([E],[T1,T2]):-
        event([E],[T1,T2]).
% Inbound complex event definitions, e.g. holdsInterval(sequence(a,b),T).
holdsInterval([Operator|Events],T):-
        operator(Operator),
        event([Operator|Events],T).
```

We omit the definitions of the partially bound formalisations here, e.g. where *T1* is bound and *T2* is free. The complete interval based Event Calculus formalization can be downloaded with the ContractLog KR available at https://sourceforge.net/projects/rbsla.

The event predicate *event([Event],[Interval])* is a meta predicate to translate instantaneous event occurrences into interval-based events: event([E],[T,T]) :- occurs(E,T). It is also used in the event algebra meta program to compute complex events from occurred raw events according to their event type definitions (see next section). The clipped predicate tests whether the event interval is not broken between the terminating time of the initiator event and the initiating time of the terminating event by a global terminator event:

```
broken(T1,Interval,T2):-
        terminates(Terminator,Interval,[T1,T2]),
        event([Terminator],[T11,T12]),
        T1<T11, T12<T2.
```

<u>Example</u>

```
occurs(a,datetime(2005,1,1,0,0,1)).
occurs(b,datetime(2005,1,1,0,0,10)).
Query: holdsInterval([a,b],Interval)?
Result: Interval = [datetime(2005,1,1,0,0,1), datetime(2005,1,1,0,0,10)]
```

### 3.4. An Event/Action Algebra based on the Interval-based Event Calculus

We now redefine the (SNOOP) event algebra operators based on the new event logics approach of the interval-based Event Calculus evolved in the previous section and treat events as occurring over an interval rather than in terms of their instantaneous detection times. The basic idea is to split the occurrence interval of a complex event into smaller intervals in which all required component events occur, which leads to the definition of event type patterns in terms of interval-based event detection conditions. In particular, these means a sequence *(A;B)* spans an interval *[A,B]*, where *A* initiates and *B* terminates the interval. For example, the intervals of a sequence *(A;B,C)* are *[A,C],* where *[A,C]* consists of *[A,B]* and *[B,C]*. According to the sequence definition a terminator for the interval *[A,B]* is *C*. Moreover, under strict interpretation repeated occurrences of events of type *A* or *B* between *[A,B]* are also interpreted as terminators, e.g. under strict interpretation only the second occurrence of the event instance *a* in an event instance sequence *{a a b}* fulfils the sequence definition *(A;B)*, whereas under non-strict interpretation both events instances of a lead to two detections of the sequence *(A;B)*. To capture both semantics (strict and non-strict) and allow local definitions of terminators (in contrast to the global terminates axioms of the classical Event Calculus) we have extended the general *holdsInterval* axioms (see section 3.3.) with an additional variable "*Terminators"* which defines a set of terminators which break the interval: *holdsInterval([E1,E2],[T1,T2],Terminators)*, where *Terminators* is a local list of terminating events $[TE_1,..,TE_N]$. The meta program (see appendix E) used to meta interpret the LP based event operators can be configured for strict and non-strict interpretation in order to automatically use the appropriate set of terminators, e.g. for a sequence definition *(A;B;C)*, i.e. *sequence(a,b,c),* under strict interpretation the terminators are *[a,b,c]* for both event intervals *[a,b]* and *[b,c]*, whereas under non-strict interpretation the terminators for *[a,b]* are *[c]* and for *[b,c]* are *[a]*. Further global and local terminating policies can be easily added to the meta program via extending the meta program or adding global terminates axioms.

Using the holdsInterval axioms the typical event algebra operators are formalized in terms of the interval-based EC formulation as follows:

- Sequence operator (;): (A;B;C) ≡
  ```
  detect(e,[T1,T3]) :-
          holdsInterval([a,b],[T1,T2],[a,b,c]),
          holdsInterval([b,c],[T2,T3],[a,b,c]),
          [T1,T2]<=[T2,T3].
  ```
- Disjunction operator (˅): (A ˅ B) ≡
  ```
  detect(e,[T1,T2]) :- holdsInterval([a],[T1,T2]).
  detect(e,[T1,T2]) :- holdsInterval([b],[T1,T2]).
  ```
- Mutual exclusive operator (xor): (A xor B) ≡
  ```
  detect(e,[T1,T2]) :-holdsInterval([a],[T1,T2]),
  not(holdsInterval([b],[T3,T4]).
  detect(e,[T1,T2]) :-holdsInterval([b],[T1,T2]),
  not(holdsInterval([a],[T3,T4]).
  ```
- Conjunction operator (˄):(A ˄ B) ≡
  ```
  detect(e,[T1,T2]):-
          holdsInterval([a],[T11,T12]),
          holdsInterval([b],[T21,T22]),
          min([T11,T12,T21,T22],T1), max([T11,T12,T21,T22],T2).
  ```
- Simultaneous operator (=): (A=B) ≡
  ```
  detect(e,[T1,T2]) :- holdsInterval([a],[T1,T2]), holdsInterval([b],[T1,T2]).
  ```
- Negation operator (¬): (¬B [A,C]) ≡
  ```
  detect(e,[T1,T2]) :- holdsInterval([a,c],[T1,T2],[b]).
  ```
- Quantification (Any): (Any(n,A)) ≡
  ```
  detect(e,[T1,T2]):-findall([T1,T2], holdsInterval([a],[T1,T2],List), modulo(List, n, [T1,T2]).
  ```
  It is possible to define the any operator as a conjunction of all events. However, this results in long event expressions. Therefore, we use the second-order predicate "findall", which derives all occurrence intervals of *a* and collects them in a list, which is the type specific event instance sequence $EIS^T$ of all events of type *A*. The module predicate iterates over the complete list and returns all elements where size of the list module n is null, i.e. size mod n = 0.
- Aperiodic Event Operator (Ap): Ap (A,[B,C]) ≡
  ```
  detect(e,[T1,T2]):-
          holdsInterval([b,c],[T11,T12],[b,c]),
          event([a],[T1,T2]),
          between([T1,T2],[T11,T12]).
  ```
- Periodic Event Operator (Per): Per (A,t,,C)
  The periodic event operator can be efficiently represented as an ECA rule (see section 4.1) which defines a periodic time interval function t which is valid during the occurrence interval of *[a,c]*.

In order to make definitions of complex events/actions in terms of event algebra operators more comfortable and remove the burden of defining all interval conditions for a particular complex event type as described above, we have implemented a meta program which implements an interval-based EC event/action algebra in terms of typical event operators with the following axioms:

```
Sequence:           sequence(E1,E2, .., En)
Disjunction:        or(E1,E2, .. , En)
Mutual exclusive:   xor(E1,E2,..,En)
Conjunction:        and(E1,E2,..,En)
Simultaneous:       concurrent(E1,E2,..,En)
Negation:           neg([ET1,..,ETn],[E1,E2])
Quantification:     any(n,E)
Aperiodic:          aperiodic(E,[E1,E2])
```

Meta-programming and meta-interpreters have their roots in the original von Neumann computer architecture where program and data treated in a uniform way and are a popular technique in logic programming [63] for representing knowledge, in particular, knowledge in the domains containing logic programs as objects of discourse. LPs representing such knowledge are called meta-programs (a.k.a. meta interpreters) and their design is referred to as meta-programming. The meta predicate *event* is used to interpret complex event definitions defined in arbitrary combinations of the above event operators:

```
event([Operator|Events],[T1,T2]):-
        operator(Operator),
        derive(Operator(Events,[T1,T12],T2,Events)).
```

, where *Operator* is bound to an event algebra operator as defined above, *Events* is the set of events used in the operator and *[T1,T2]* is the occurrence interval. The meta program meta interprets the complex event/action definitions and derives all instances of the defined (complex) event/action type by means of the meta implementations of the event algebra operators which are based on the interval-based Event Calculus. For example the meta implementation of the sequence operator in the meta program is as follows:

```
sequence([E],[T1,T2],T2,Terminators):-
        holdsInterval([E],[T1,T2]).
sequence([E1|Rest],[T1,T2],Tend, Terminators):-
        head(E2,Rest),
        holdsInterval([E1,E2],[T1,T2],Terminators),
        sequence(Rest,[T2,T3],Tend,Terminators),
        lessequ([T1,T2],[T2,T3]).
```

The meta rule recursively iterates over all component intervals and tests the correct order of the components' occurrence interval as well as its terminating conditions given by the set of terminators from the variable Terminators until the last terminating event is reached. The implementation of the sequence operator in the interval-based EC meta program can be found in appendix E.

<u>Examples:</u>

```
Event Pattern is A ; (B ; C):      (Sequence)
detect(e,T) :- event(sequence(a, sequence(b,c)), T).
```

Accordingly, under strict interpretation the EIS *{a b c}* does detect the complex event, *{b a c}* does not detect the event, *{a c b}* does not detect the event, *{a b a c}* does not detect event and *{a b b a b c}* does detect one event.

```
Event Pattern is A (A, B,C):       (Aperiodic)
detect(e,T):- event(aperiodic(b,[a,c]),T).
```

Here the EIS *{a b c}* does detect the event, *{a b b c}* does detect the event, but *{a c b b}* does not detect the event.

Event Pattern is A ; (¬B [A,C]) ; A ∨ B (Combination)
detect(e,T) :- event(sequence(a,neg(b,[a,c]), or(a,b)),T).

### 3.5. Event Selection and Consumption via ID-based Knowledge Updates

In order to reuse detected complex events in rules, e.g. ECA rules, or in other event type pattern definitions, i.e. for the detection of other complex events they, need to be remembered until they are consumed. Furthermore, the contributing component events of a detected complex event should be consumed after detection of a complex event. This can be achieved via update primitives, which allow adding or removing knowledge from the KB. We have implemented a set of expressive ID-based update primitives. In fact they are more expressive than the simple assert/retract primitives found in typical Prolog interpreters and derivates and allow transactional as well as bulk updates of knowledge including updating of facts and rules. They enable arbitrary knowledge updates, e.g. adding (*add*) /removing (*remove*) rules or complete rule sets including integration of knowledge from external sources. Each update has a unique ID (key), with which it is asserted into the KB. Based on this id, local reasoning, modularization of rule sets, prioritization of conflicting rule sets (modules) (e.g., *overrides(module id1,module id2)*) and bulk retracts removing sets of facts/rules via its ID becomes possible, e.g.:

```
add("./examples/test/test.prova").            % add an external script
add(id1,"r(1):-f(1). f(1).").       % add rule "r(1):-f(1)." and fact "f(1)."
add(id2,"r(X):-f(X).").             % add rule "r(X):-f(X)."
add(id3,"r(_0):-f(_0), g(_0). f(_0). g(_1).",[1,2]). % update with variable/
                                                % object place holders _X.

remove(id1).                                  % remove all updates with id
remove("./examples/test/test.prova").         % remove external update

transaction(remove(...)). % transactional remove with internal integrity/
                          % test case test
transaction(add(...)). % transaction update
transaction(add(...),"testcase1.prova"). % transaction update with explicit
                                          test case test.
commit(id5).                   % commit transaction with ID id5
rollback(id5).                 % rollback transaction with ID id5.
```

The transactional update primitives provide support for integrity constraints and test cases. If an update violates the integrity constraints or test cases defined in the LP or in an external test case the update/remove transaction is automatically rolled back.

An integrity constraint expresses a condition which must always hold. We support four basic types of integrity constraints:

1. **Not-constraints** which express that none of the stated conclusions should be drawn: integrity( not( $p_1(\dots)$, .. , $p_n(\dots)$)).
2. **Xor-constraints** which express that the stated conclusions are mutual exclusive, i.e. should not be drawn at the same time: integrity(xor($p_1(\dots)$,..,$p_n(\dots)$)).
3. **Or-constraints** which express that at least one of the stated conclusions should be drawn: integrity(or($p_1(\dots)$,..,$p_n(\dots)$)).
4. **And-constraints** which express that all of the stated conclusion should be drawn: integrity (and($p_1(\dots)$, .. , $p_n(\dots)$)).

Literals (predicates) might be also negated. Integrity constraints might be conditional, i.e. stated as integrity rules of the form: integrity(…):- $b_1(\dots)$..$b_m(\dots)$ . We have implemented a meta program (see appendix F) which is used to test (meta interpret) the defined integrity constraints:

1. **testIntegrity()** iterates over all integrity constraints in the knowledge base and tests them based on the actual facts and rules in the knowledge base.
2. **testIntegrity(Literal)** tests the integrity of the literal, i.e. it makes a test of the (hypothetically) added/removed literal, which might be a fact or the head of a rule.

<u>Example:</u>

```
neg(p(x)).          %fact
integrity(xor(p(x), neg(p(x)))). % mutual exclusive integrity constraint
testIntegrity(p(x))?              %test integrity
```

The example defines an explicitly negated predicate "p(x)" and an integrity constraint, which states that p(x) and the negation of p(x) are mutual exclusive. Accordingly, the hypothetical test "testIntegrity(p(x))" fails. Test cases generalize the concept of integrity constraints and in addition might provide further optional meta test rules and test facts (assertions) which are used during testing.
For a more detailed description of the syntax and semantics of the labelled and unitized logic in ContractLog and the OID-based update functions as well as the verification, validation and integrity (V&V&I) approach in the ContractLog KR, which underpins ECA-LP and ECA-RuleML, see [64].

We use these update primitives and update formalisms to add detected events as new transient occurrence facts to the KB and consume the events which have contributed to the detection of the complex event.

<u>Example</u>

```
detect(e,T):-
        event(sequence(a,b),T), % detection condition for the event e
        add(eis(e), "occurs(e,_0).", [T]), % add e with key eis(e)
        consume(eis(a)), consume(eis(b)). % consume all a and b
events
```

If the detection conditions for the complex event *e* are fulfilled, the occurrence of the detected event *e* is added to the KB with the key *eis(e)*. Then all events that belong to the type specific event instance sequences of type *a* and type *b* are consumed using their ids *eis(a)* resp. *eis(b)*. Different con-sumption policies are supported for consume, such as remove all events which belong to a particular type specific *eis* or remove the first resp. the last event in the eis. If no consume predicate is specified in a detection rule, all events can be reused, e.g. for the detection of other complex events.

As we have shown, due to the style of event definition in terms of their occurrence intervals the problems with the unintended semantics of certain event operators in the active database algebras can be solved. Moreover the proposed event algebra is based on a sound and complete, declarative logical semantics stemming from the typical LP semantics for derivation rules and the meta LP formalization of the Event Calculus which has been extended to an interval based Event Calculus. This is different to the event algebras and event processing systems in the active database domain which typically only have a procedural operational semantics. The event definitions can be easily combined with other logical formalisms in the context of logic programming. This clearly leads to higher levels of expressiveness than traditional event algebras have. The task of the event detection mechanism is supported directly by the resolution algorithm of the inference engine, which, due to the formal, mathematical basis, enables traceability and validation of the detected events. The approach is suitable for applications where the ability to reason formally about the system is crucial for describing reliable real-world decision logic and trigger rule based reactions in response to occurred events. Moreover, the proposed formalization fulfils typical criteria for good language design [65] including minimality, symmetry and orthogonality. Minimality means that the language provides only a small set of needed language constructs, i.e., the same meaning cannot be expressed by different language constructs. Symmetry is fulfilled if the same language construct always expresses the same semantics regardless of the context it is used in. Orthogonality permits every meaningful combination of a language constructs to be applicable. The formalization in terms of an interval-based Event Calculus provides an intuitive readability of the associations between event occurrences and the time intervals they initiate or terminate. Although we have mainly focused on complex event definitions in this section it should be noted that the event algebra operators can be equally applied on complex action definitions, e.g., to define a sequence of actions.

In the next section, we describe two approaches of event processing, namely active pull-based ECA rules and push-based event notifications, and relate them to the interval-based event/action algebra defined in this section.

## 4. Event Processing in Logic Programming Environments based on ECA rules

We distinguish two basic approaches in event processing: *pull-based* and *push-based* models. The former actively monitors the environment for occurred events or pulls events from a central storage and triggers reactions by means of ECA rules, whereas the latter uses an external component to push events to the event system, i.e. the active event memory which then processes the event facts. In this section, we elaborate on these approaches in the context of Logic Programming and KR based event/action logics and related them to our interval-based EC event algebra.

Event-Condition-Action rules (ECA rules) inherently have an active forward-directed operational semantics, i.e. the condition of a rule is evaluated, whenever a particular event is detected and when the condition is satisfied a specified action is executed. Active database systems or production rule systems typically implement this ECA paradigm purely procedural or based on forward reasoning, e.g. via the RETE algorithm. However, among many other disadvantages procedural or forward-reasoning solutions have no clear formal semantics and amount to imperative programming. In contrast, backward-reasoning rule systems, as used in logic programming, have a goal-driven logical semantics which needs an initial query to start a deductive refutation attempt based on the concept of resolution and unification. Logic programming has been proven to be one of the most successful approaches in the area of declarative programming. This prompts us to attempt a tighter integration of ECA rules into logic programming and backward-reasoning rule engines. The key to success is to implement each constituent part of an ECA rule as a derivation rule and repeatedly query those rules according to the forward-directed operational semantics of the ECA paradigm, i.e. first query the derivation rule describing the event part; if it succeeds then the condition part is evaluated via querying the condition rule and finally the action is executed.

We define a reactive rule as an extended ECA rule represented as a 6-ary fact: ***eca(T,E,C,A,P,EL)***, where *T* (time), *E* (event), *C* (condition), *A* (action), *P* (post condition), *EL*(se) are complex terms, which are interpreted by the ECA processor as queries on derivation rules which implement the respective functionality of each of the ECA rules' parts.

- The *time part* of an ECA rule defines a pre-condition (an explicitly stated temporal event) which specifies a specific point in time at which the ECA rule should be processed by the ECA processor, either absolutely (e.g., "at 1 o'clock on the 1st of May 2006), relatively (e.g., 1 minute after event X was detected) or periodically ("e.g., "every 10 seconds"). In other words, it defines a validity period or monitoring schedule for the ECA rule in order to address real-world requirements, such as event detection costs (i.e. leads to reduce monitoring costs and increased scalability).
- The *post-condition* is evaluated after the action has been fired. It might be used to prevent backtracking from different variable bindings carrying the context information from the event or condition part via setting a cut "!", i.e. execute an action only once for one binding. An example might be a list of flights which is bound in the condition part of the ECA rule, where the first permissible flight is booked in the action part, i.e. the processor backtracks as long as the action does not succeed, otherwise the cut applies and the ECA rule finishes. Furthermore, the post condition might be used to do integrity tests based on integrity constraints and test cases [64] which must hold after the action is executed, e.g., an internal transactional knowledge update in the action part which violates the post condition is automatically rolled back. Note that side effects such as updates in external systems via procedural attachments can not be rolled back unless the external component supports transaction mechanisms.

- The *else part* defines an alternative action which is execute in case the ECA rule fails, e.g. to specify a default action or trigger some failure handling action.

To illustrate this operational ECA semantics of the ECA processor, consider an ECA rule which states that "*every 10 seconds it is checked (time) whether there is an incoming request by a customer to book a flight to a certain destination (event). Whenever this event is detected, a database look-up selects a list of all flights to this destination (condition) and tries to book the first flight (action). In case this action fails, the system will backtrack and book the next flight in the list otherwise it succeeds (post-condition cut) sending a "flight booked" notification If no flight can be found to this destination, i.e. the condition fails, the else action is triggered, sending a "booked up" notification back to the customer.*" This is formalized as a LP (Prolog related syntax) as follows:

```
eca(
        every10Sec(),
        detect(request(Customer, Destination),T),
        find(Destination, Flight),
        book(Customer, Flight),
        !,
        notify(Customer, bookedUp(Destination)
).

% time derivation rule
every10Sec() :-  sysTime(T), interval( timespan(0,0,0,10),T).

% event derivation rule
detect(request(Customer, FlightDestination),T):-
        occurs(request(Customer,FlightDestination),T),
        consume(request(Customer,FlightDestination)).

% condition derivation rule
find(Destination,Flight) :-
         on_exception(java.sql.SQLException,on_db_exception()),
        dbopen("flights",DB),
        sql_select(DB,"flights", [flight, Flight], [where,
"dest=Destination"]).

% action derivation rule
book(Cust, Flight) :- flight.BookingSystem.book(Flight, Cust),
                        notify(Cust,flightBooked(Flight)).

% alternative action derivation rule
notify(Customer, Message):-
                        sendMessage(Customer, Message).
```

The example includes possible backtracking to different variable bindings. If the action succeeds further backtracking is prevented by the post-conditional cut. If no flight can be found for the customer request, the else action is executed which notifies the customer about this. The condition derivation rule accesses an external data source via an SQL query. Each part in the ECA rule specifies a query on the globally defined derivation rules, implementing the respective functionality. Accordingly, the basic declarative ECA rule syntax remains compact and enables reusing the separated, declarative rule specifications several times within different ECA rules. Note that the ECA rules and the derivation rules supporting the ECA rules have a declarative LP semantics, i.e. the ECA rules (which are represented as facts) and derivation rules are defined globally and their order is not important. This is congruent to the declarative semantics of traditional logic programs. The ECA rules are meta interpreted by the operational semantics of the ECA processor. The LP inference engine[5] is used to compute variable bindings, (non-monotonic) negations (strong negation in the sense of explicit classical negated knowledge as well as non-monotonic weak negation in the sense of negation as failure) and other logical connectives including automated rule chaining (based on resolution) to build-up complex decision logics in terms of derivation rules.

ECA parts might be blank, i.e. always true, stated with "_", e.g., eca(time(), event(...), _, action(...),_,_) or completely omitted, e.g. eca(e(...),c(...),a(...)). This leads to specific types of reactive rules, e.g. common ECA rules (ECA: eca(event(),condition(),action()) ), production rules (CA: eca(condition(),action()) ), extended ECA rules with post condition (ECAP: eca(event(),condition(),action(), postcondition()) ). Semantically a blank or omitted rule part is automatically assigned true, hence leading to the final success of the ECA rule (which is semantically interpreted as a top goal) if no other rule parts fail. The rule engine which we use supports procedural attachments on external Java implementations (e.g. flight.BookingSystem.book(Flight, Customer) calls the method "book" of class "flight.BookingSystem") and data integration from external data sources as well as communication built-ins and messaging capabilities (cf. section 4.2). The tight combination of ECA rules and derivation rules within the framework of logic programming paves the way to (re-)use various other useful logical formalisms, such as defeasible reasoning for conflict resolution, event logics for temporal event calculations within ECA rules and in general, it relates ECA rules to other rule types such as business rules, integrity constraints or normative rules. For the rest of this section we will first elaborate on the implementation of our ECA processor within backward-reasoning LP environments and then illustrate the beneficing or even necessity to include event logics, in particular the Event Calculus, into reactive rules.

[5] Depending on the supported class of LP (e.g. normal LP, extended LP, disjunctive LP) and the logical semantics (e.g. well-founded, stable models, answer set)

### The Logic Programming ECA Processor

Figure 1 illustrates the process of frequently querying the constituent ECA parts according to the forward-directed operational semantics of the ECA paradigm.

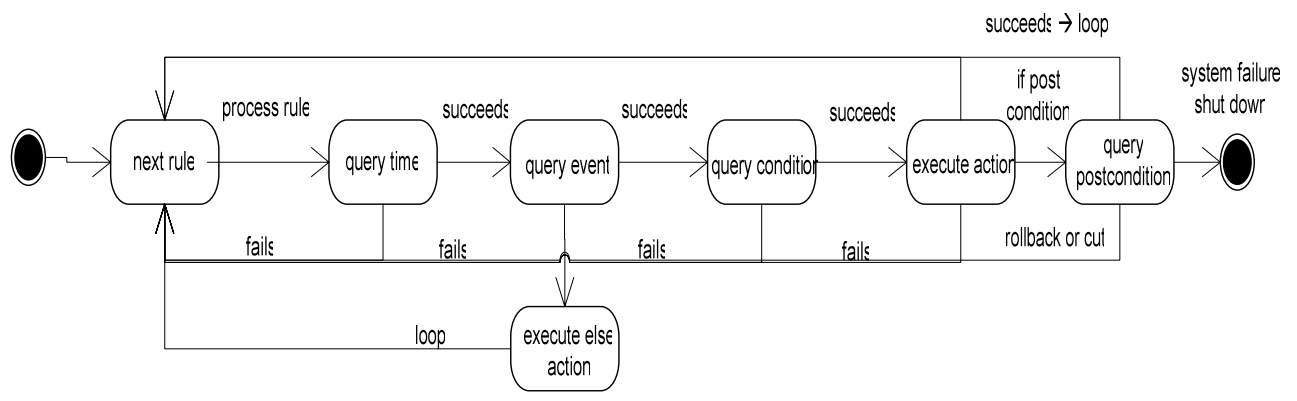

**Figure 1:** Forward-directed execution model of ECA rules

This task is solved by a **demon** (implemented within the ECA processor), which is a kind of adapter that frequently issues queries on the ECA rules. It queries the KB for ECA rules *eca(T,E,C,A,P,EL)?*, adds the resulting ECA rules to the active KB, which is a kind of volatile storage for reactive rules, and evaluates the ECA rules one after another via using the complex terms defined within the ECA rule as queries on the associated derivation rules. Obviously, strictly sequential processing (querying) of the ECA rules would lead to large processing delays, e.g. due to long proof trees or external queries such as complex SQL statements or Internet queries. Therefore, we have extended the demon with a thread pool and evaluate each ECA rule in a thread, i.e. if the time part of an ECA rule succeeds a free thread is requested from the thread pool and the succeeding ECA parts (event, condition, action, post condition) are processed within this thread. In parallel the demon can proceed with the next ECA rule, which enables concurrent execution of ECA rules. Moreover, time outs can be set to overcome termination problems, e.g. waiting for answers from external systems. Figure 2 illustrates the extended operational semantics of the ECA processor.

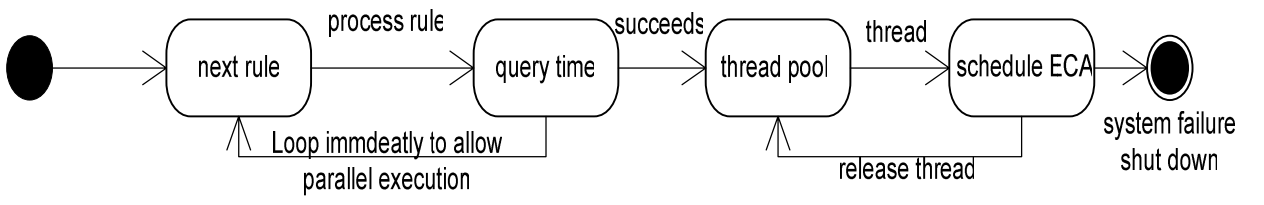

**Figure 2:** Concurrent execution of ECA rule using threads

### Dynamic Active ECA Rules

ECA rules do not only affect the outside world (via procedural attachments) but often also the internal knowledge system in terms of **state changes** and **updates**. In section 3.5 we have elaborated on a set of id-based expressive update primitives. The updates are persistent and not just hypothetical calculi and can be applied in transactional style. They enable arbitrary knowledge updates, e.g. adding (*update*) /removing (*remove*) rules or complete rule sets which might possibly specified in external files. These update primitive might be used in the action part of an ECA rule in order to update and evolve the knowledge system as a reaction on occurred events. However, there might be problems when using updates in a **transactional style**, i.e. more than one update in a sequence must be completely executed and in style of **active rules**, where an update itself is seen as an event which again might entail another update. The standard operational semantics of ECA rule interpreters domain does not address these topics. In particular it gives no answers how update primitives should interact with other logical operators such as disjunction and negation and how the order of updates can be logically treated, e.g. depending on the order of updates under strict procedural semantics the outcome of an execution might vary so that a unique outcome can not always guaranteed, which is called a confluence problem. Another problem is if a particular update in a transaction (sequence of updates) fails - shall the system proceed with the next update, simply abort, or rollback all previous updates? Using active rules can lead to loops due to cyclic dependencies between updates which are conditional to each other and therefore induce termination problems.

Several solutions to these problems have been proposed in particular in the area of active and deductive databases applying special logics such as transaction logics, dynamic logics, process logics or applying special techniques for checking termination and confluence properties as well as approaches for rewriting update programs consisting of active rules into deductive rules and applying confluent semantics such as well-founded semantics. Other solutions are more software engineering related and make use of techniques from the database world with its ACID properties, locks and its commit protocols e.g. 2-phase commits or annotate events/updates with time stamps to explicitly define the order of events/updates. However, most of these approaches need extensive extensions to the standard rule engines and LP logics, e.g. new logic types such as transaction logics or additional special systems annotating events with e.g. with time stamps. Applying declarative, confluent semantics (e.g. well-founded semantics) solves some of the mentioned problems such as termination (for function free queries) and to some extend confluence - however needs heavy transformation approaches [34], e.g. from execution models for active rules to derivation rule LPs. However, this creates other problems, e.g. parallel execution of ECA rules is forbidden. In a nutshell, most of the approaches are not applicable in standard backward-reasoning rule engines without heavy reimplementation and extensions and would contradict our design goals of reusing the existing functionalities of LPs.

Therefore, we apply a test-driven approach based on integrity constraints and test cases [64] to safeguard update transactions and apply rollbacks (*rollback*) in case of failures and conflicts. The idea is that after the execution of the complete transaction which might be a complex action consisting e.g. of a sequence of updates which transits the initial state of the program/kb in a sequence of intermediate evolving states to the final knowledge state, the integrity constraints must still hold and the defined test cases must succeed for the final logic program state. If this is not the case the transaction is rolled back (see section 3.5). The meta program to test integrity constraints can be found in appendix F. The complete implementation is provided with the ContractLog KR and you will find further information in [64].

**Relating the Event Calculus to ECA Rules**

The interval-based EC event algebra defined in section 3 can be easily integrated into ECA rules via querying the detection rules defining the detection conditions for complex events, e.g.:

Example

```
eca(everySecond(), detect(e,T),_,fireAction()).
detect(e,T):-
    event(sequence(a,b),T),
    consume(eis(a)), consume(eis(b)).
```

Moreover, the EC is also useful to capture the actual context in which processing of an ECA rule is done, in particular in the condition part to compute the actual temporal or state characteristics under a long-term retrospective view as a computation of persistent earlier events (event history). For example an ECA rule might define that *(the state) escalation level 1 is triggered (action) in case a service "s" is detected to be unavailable via pinging/monitoring the service (event) every minute (time), except we are currently in a maintenance state (condition)*. This can be formalized as an ECA rule:

```
eca(
    everyMinute(),
    detect(unavailable(s),T),
    not(holdsAt(maintenance(s),T)),
    add("",happens(unavailable(s),_0),[T]) ).
```

, i.e., in the condition part it is evaluated whether the state *maintenance* for the service *s* holds at the time of detection of the *unavailable* event or not. In case it does not hold, i.e. the condition succeeds, the detected transient event *unavailable(s)* is added to the KB as a non-transient event in terms of a EC "happens" fact, in order to initiate escalation level 1:

```
initiates(unavailable(s), escl(1),T).
```

Additionally, we might define that the *unavailable* event can not be detected again as long as the state *escl(1)* has not been terminated and accordingly the ECA rule will not fire again in this state:

```
detect(unavailable(s),T) :- sysTime(CT), not(holdsAt(escl(1),CT)), ....
```

This exactly captures the intended semantics of this reactive rule. It should be noted that the event calculus state processing capabilities can be also used to apply sophisticated event calculus based post-conditional test after internal update actions which change certain properties of the knowledge system and hence must hold at the succeeding state after the update action.

In summary, the EC can be effectively used to model the effects of events and actions on the knowledge states and describe sophisticated state transitions akin to state machines. In contrast to the original use of ECA rules in active database management systems to trigger timely response when situations of interest occur which are detected by volatile vanishing events, the integration of event logics KR formalisms adds temporal reasoning on the effects of non-transient, happened (or planned) events on the knowledge system, i.e. enable "state tracking". They allow building complex decision logics upon, based on a sound declarative semantics of the interval-based Event Calculus formalization as opposed to the database implementations which only have an operational semantics. As a result, due to the formal semantics which assigns truth values as a result of reasoning or execution process in each constituent part of an ECA rule (see ECA-LP declarative semantics [35]) the derived conclusions and triggered actions become traceable and verifiable, which is a crucial necessity to build reliable real world production systems upon, such as business rule systems.

## 5. ECA-RuleML: A declarative ECA Rule Markup Language based on RuleML

Existing work on ECA rule markup languages is still very much in its initial phase. RuleML (Rule Markup Language) is a standardization initiative with the goal of creating an open, producer-independent XML/RDF based web language for rules, foresees reaction rules but the syntax has not been specified yet. However, it already provides a rich syntax for derivation rules and supports different logic classes such as datalog, hornlog (with naf), extended (with neg) disjunctive, FOL, which makes it an adequate candidate to base our ECA rule markup language (ECA-RuleML) upon. Our design goals are to stay as close as possible to the RuleML standard, reuse the existing language constructs and fulfill typical criteria of good language design such as minimality, homogeneity, symmetry and orthogonality. [65] In particular, we try to keep the set of new language constructs as small as possible and give these constructs a plain declarative semantics. We follow the design principle of RuleML and define the new constructs within separated modules which are added to RuleML as additional ECA-RuleML layers, i.e. it adds additional expressiveness and modeling power to RuleML for the serialization of reactive rules (ECA rules) including event definitions and action language constructs, event notifications, event logics (Event Calculus) and event algebras (complex event pattern definitions) in XML. We use XML Schema group definitions to define language constructs which belong together in a group. This approach is easily extensible, i.e. it is easy to add new constructs, e.g. add a new event operator to the group of event operators so that it can be automatically used wherever the operator group is used, e.g. in an ECA rule or in the holdsInterval atom. We first describe relevant constructs in RuleML and then introduce the additional layers in ECA-RuleML.

### 5.1. Fundamentals of RuleML

An overview over the complete RuleML structure can be found in appendix C. Here, we focus on the RuleML horn logic layer extended with negation and equality. The building blocks are: [66]

**Predicates** (atoms) are n-ary relations defined as an <Atom> element in RuleML. The main terms within an atom are **var**iables <Var> to be instantiated by ground values when the rules are applied, **ind**ividual constants <Ind>, **data** values <Data> and **c**omplex **term**s <Cterm>.

**Derivation Rules** (<Implies>) consist of a body part (<body>) with one or more conditions (atoms) connected via <And> or <Or> and possibly negated by <Neg> which represents classical negation or <Naf> which represents negation as failure and a conclusion (<head>) which is derived from existing other rules or facts applied in a forward or backward manner.

**Facts** are deemed to be always true and are stated as atoms: <Atom>

**Queries** <Queries> can either be proved backward as top-down goals or forward via bottom-up processing. Several goals might be connected within a query and negated.

Besides facts, derivation rules and queries RuleML defines further rule types such as **integrity constraints** and **transformation rules**.

### 5.2. The ECA RuleML Layer

We base the ECA Rule Markup Language (ECA-RuleML) on the Horn logic layer of RuleML (hornlog). We extend this layer within the "hornlog2eca" layer in order to add adequate expressiveness for the superimposed ECA and Event Calculus syntax which is implemented in the "eca" and "ec" module and integrated into the "ECA" and "event_calculus" layer (cf. fig. 3). The redefinitions and module extensions within the hornlog2eca layer relate to strong negation (Neg) and negation as failure (Naf), equality, procedural attachments, update primitives and input/output mode declarations, to name some.

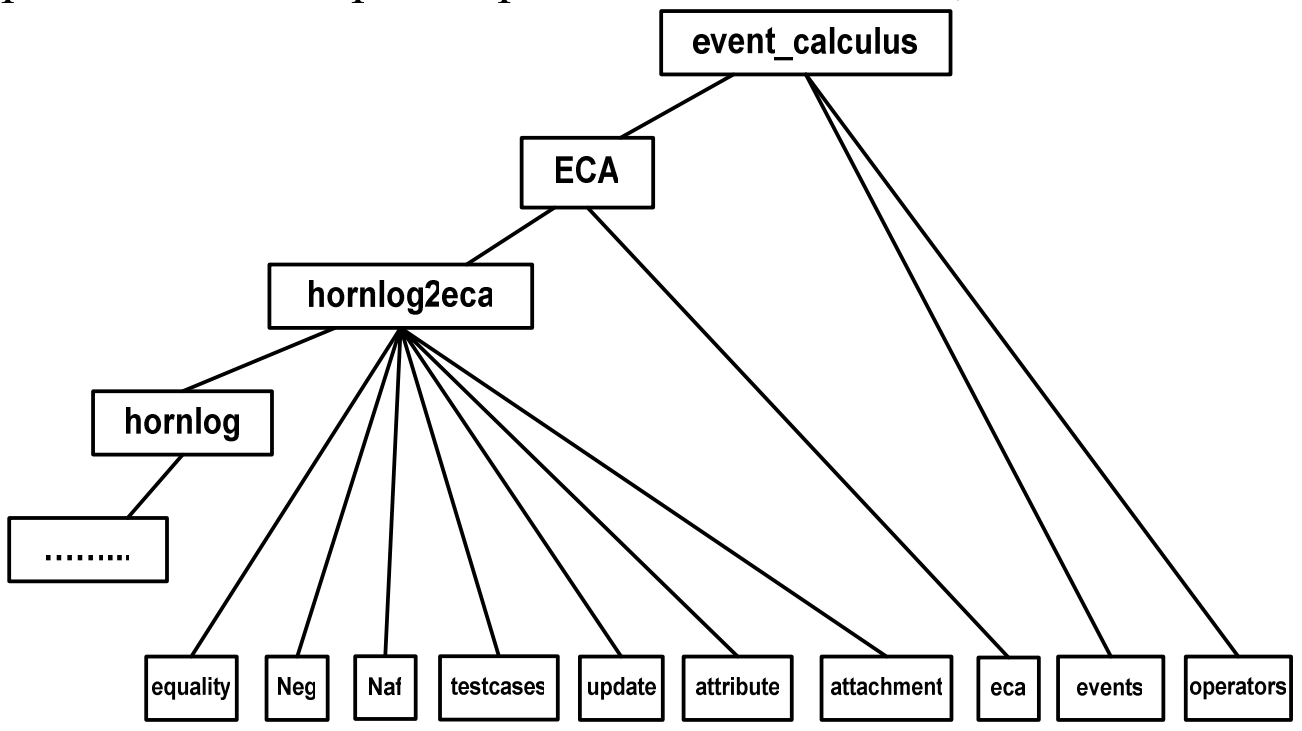

Figure 3: ECA-RuleML structure

In the following we describe the main language constructs coming with ECA-RuleML. We specify the ECA syntax by means of extended BNF (EBNF), i.e. alternatives are separated by vertical bars (|); zero to one occurrences are written in square brackets ([]) and zero to many occurrences in braces ({}).

**hornlog2eca Layer**

The hornlog2eca layer acts as an intermediate between RuleML and the ECA language. It redefines several RuleML constructs and adds several extensions to RuleML such as test cases to validate logic program specifications, update primitives to integrate external knowledge (modules) and add/remove internal knowledge or type and mode declarations to define polymorphic types and input/output definitions.

**Modes** are states of instantiation of the predicate described by mode declarations, i.e. declarations of the intended input-output constellations of the predicate terms with the following semantics:

"+" The term is intended to be input
"-" The term is intended to be output
"?" The term is undefined/arbitrary (input or output)

Modes are frequently used in inductive logic programming (ILP) to reduce the space of clauses actually searched, i.e. to narrow the hypothesis space of program clauses structured by theta-subsumption generality ordering (a.k.a. refinement graph). The idea is that a rule system is written in such a way that in a computed answer the output variables are contained in the input variables of a query or goal. Thus in a certain way they reflect the data flow of a rule set. In ECA-RuleML we define modes with an optional attribute *@mode* which is added to terms, e.g. <Var mode="-">X</Var>, i.e. the variable X is an output variable. By default the mode is undefined "?".

**Types** typically define a type relation "t:r", denoting that term t has a type r . Types in ECA-RuleML can be assigned to terms using a *@type* attribute. ECA-RuleML supports primitive built-in data types such as String, Integer and XML Schema built-in data types such as *xs:dateTime* as well as external type systems such as Java class hierarchies (fully qualified Java class names), e.g. *type="java.lang.Integer"* or Semantic Web taxonomies based on RDFS or OWL, e.g. *type="rbsla_Provider"*, where *rbsla* is the namespace prefix and *Provider* is the concept class.

Example:

<Var **type=**"***java.lang.Integer***">1234</Var>
<Var **type="*rbsla_Provider***">Service Provider</Var>

In the context of event processing and action execution an important extension to RuleML are **procedural attachments**. Procedural attachments are in particular relevant for actively accessing / monitor external systems and detecting events in a pull mode. They allow receiving/integrating external information, affecting the outside world or delegating computation-intensive tasks to optimized procedural code (e.g. Java methods). Hence, they are a crucial extension of the pure logical inferences used in logic programming. We model a procedural attachment within a complex term (<Cterm>), where the attachment (<Attachment>) is used instead of the Cterms' constructor (<Ctor>), i.e. the attachment calls an external piece of code which takes the provided input data stated in the Cterm arguments and constructs a result which might be a true/false value or one or more result objects of a particular type. These objects might be bound to variables for further use in the refutation/derivation attempt via equality relations.

<u>Procedural Attachment</u>
*Cterm* ::= [*oid*,] *op* | *Ctor* | *Attachment*, {*slot*,} [resl,] {*arg* | *Ind* | *Data* | *Skolem* | *Var* | *Reify* | *Cterm* | *Plex*}, [*repo*], {*slot*}, [*resl*]
*Attachment* ::= [*oid*,] *Ind* | *Var* | *Cterm*, *Ind*

<u>Example:</u> call "System.out.print("Hello!")"

```
<Cterm>
   <Attachment>
      <Ind type="java class"> System.out </Ind>
      <Ind type="java method"> print </Ind>
   </Attachment>
   <Ind type="java.lang.String"> Hello! </Ind>
<Cterm>
```

<u>Example:</u> X = rbsla.utils.Math.add(1,2) add two integers and bind result to the variable X. The method add has the modes add(+,+) indicating that both arguments are input terms for the method.

```
<Equal>
      <side><Var type="java.lang.Integer>X</Var></side>
      <side><Cterm>
          <Attachment>
                <Ind>rbsla.utils.Math</Ind>
                <Ind>add</Ind>
          </Attachment>
          <Ind type="java.lang.Integer" mode="+">1</Ind>
          <Ind type="java.lang.Integer" mode="+">2</Ind>
      </side>
</Equal>
```

Another important extension are **ID based update primitives**, which allow dynamically adding or removing internal and external knowledge and evolve the knowledge system at runtime.

<u>ID based update constructs</u>
*Assert* ::= *content* | *And*
*Retract* ::= *content* | *And*
*RetractAll* ::= *content* | *And*

The update primitives are defined on the level of atoms and can be used within rules. This is different from the assert construct in RuleML which is a KQML like performative defined on the top level of RuleML. An attribute @safty states whether the update should be performed in transactional mode, i.e. it might be rolled back in case of failures. External knowledge files which should be asserted can be referenced within the oid tag defined under And, e.g.:

```
<Assert>
   <And>
      <oid>
         <Ind>./examples/ContractLog/RBSLA/math.rbsla</Ind>
      </oid>
   </And>
</Assert>
```

<u>Example:</u> eca( …, add(id1,"f(). p():-f()."")).

```
<ECA>
   …
   <action>
      <Assert safety="transactional">
         <And>
            <oid><Ind>id1</Ind></oid>
            <Atom><Rel>f</Rel></Atom>
            <Implies>
               <Atom><Rel>f</Rel></Atom>
               <Atom><Rel>p</Rel></Atom>
            </Implies>
         </And>
      </Assert>
   </action>
</ECA>
```

The example shows an internal knowledge updated with the id "id1" adding a fact and a rule as an action of an ECA rule. Due to the transactional mode, the update should be rolled back, if it is only partially performed or violates any integrity constraint/test case.

## <u>ECA Layer</u>

The ECA layer respectively the eca module defines the syntax for ECA rules:

<u>ECA Rule</u>
*ECA* ::= [*oid*,] [*time*,] [*event*,] [*condition*,] *action* [,*postcondition*] [,else]
*time* ::= *Naf* | *Neg* | *Cterm*
*event* ::= *Naf* | *Neg* | *Cterm* | *Sequence* | *Or* | *Xor* | *And* | *Concurrent* | *Not* | *Any* | *Aperiodic* | *Periodic*
*condition ::= Naf* | *Neg* | *Cterm*
*action* ::= *Cterm* | *Assert* | *Retract* | *RetractAll*
*postcondition* ::= *Naf* | *Neg* | *Cterm*
*else* ::= *Cterm* | *Assert* | *Retract* | *RetractAll*

*Naf* ::= [*oid*,] weak | *Atom* | *Cterm*
*Neg* ::= [*oid*,] strong | *Atom* | *Equal* | *Cterm*

According to the definition in section 4 an ECA rule (<ECA>) consists of six parts and an optional object id (<oid>, its rule label). Like in RuleML, the syntax is a striped syntax with two kinds of tags: method-like role tags, such as <time>, <event>, <action>, which start with a lower-case letter and class-like type tags such as <Neg>,<Cterm> which start with an upper-case letter in order to be compatible with RDF. At least the <action> within an ECA rule must be specified, which leads to a rule which repeatedly executes an action. Other examples are Condition-Action rules or Event-Action rules. The ECA parts are modeled as complex terms which might be negated, e.g. a strongly negated (<Neg>) time part means, that the rule will be further evaluated when the query on a time function implemented in a derivation rule or an external procedural code (called via an procedural attachment) fails, e.g., when we are not within a specified time period or a particular time point does not hold. Negation as failure (<Naf>) is used to handle unknown knowledge, e.g. a default negated event part means that the rule fires when the system can not detect the event, maybe because it does not have enough information or the query on an external system returns no answer. Note that the syntax abstracts away from any procedural or control aspects of the underlying inference engine, in particular from the inference direction (backward vs. forward). The Cterms can be either interpreted as queries on derivation rules starting a refutation attempt (backward reasoning) as described in the section 4 or they can be interpreted as intended results of a bottom-.up derivation process triggered by a fact change (forward reasoning). Both cases **construct** some kind of knowledge such as a newly detected event, a derived condition etc. Therefore, complex terms are a suitable language concept in RuleML to capture this semantics and to enable reification style of nested definitions. Moreover, Boolean-valued procedural attachments defined within Cterms can be used directly within an ECA rule constructing a true / false answer for a particular ECA part based on a method in an externally defined code.

Example

```
<ECA>
   <time><Cterm><Ctor>everyMinute</Ctor><Var>T</Var></Cterm>
   <event><Cterm><Attachment>
        <Ind type="java class"> WebService </Ind><Ind> ping </Ind>
   </Attachment><Ind>http://www.rbsla.de</Ind</Cterm></event>
   <action><Assert><And><oid><Ind>eis(unavailable)</Ind></oid>

<Atom><Rel>occurs</Rel><Ind>unavailable</Ind><Var>T</Var></Atom>
     </action>
</ECA>
```

This ECA rule every minute pings a service via a procedural attachment on the static method *ping* of the Java Web*Service* class and *updates* the KB with the fact that the event *unavailable occurred* at time *T*. Formalized as a LP this can be expressed as follows:

```
eca(
   everyMinute(T),
    rbsla.utils.WebService.ping(http://www.rbsla.de), _,
   add(eis(unavailable),"occurs(unavailable(http://www.rbsla.de),_0)",[T])
).
```

## Event Calculus Layer

For the rest of this section we will turn our focus on the serialization of the event calculus axioms and the event algebra operators.

Event Calculus

```
event ::=  Ind | Var | Cterm
fluent ::= Ind | Var |Cterm
parameter ::= Ind | Var |Cterm
time :: =  Ind | Var | Cterm
interval::=Interval | Plex | Var

Happens ::= [oid,] event | Ind | Var | Cterm ,  time |  Ind | Var | Cterm
Planned ::= [oid,] event | Ind | Var | Cterm,  time |  Ind | Var | Cterm
Occurs ::= [oid,], event | Ind | Var | Cterm, interval | Interval | Plex | Var
Initially ::= [oid,] fluent | Ind | Var | Cterm
Initiates ::= [oid,] event | Ind | Var | Cterm, fluent | Ind | Var | Cterm,
          time | Ind | Var | Cterm
Terminates ::= [oid,] event | Ind | Var | Cterm, fluent | Ind | Var | Cterm |
          interval | Interval, time | Ind | Var | Cterm | interval | Interval
HoldsAt ::= [oid,] fluent | Ind | Var | Cterm , time | Ind | Var | Cterm
ValueAt ::= [oid,] parameter | Ind | Var | Cterm, time | Ind | Var | Cterm, Ind |
          Var | Cterm | Data
HoldsInterval ::= [oid,] interval | Interval | Plex | operator | Sequence | Or |
               Xor | And | Concurrent | Not | Any | Aperiodic |  Periodic |
               Cterm , interval | Interval | Plex | Var
Interval::= [oid,] event | time | Ind | Var | Cterm | operator | Sequence | Or |
          Xor | And | Concurrent | Not | Any | Aperiodic |  Periodic , event|
          time | Ind | Var | Cterm | operator | Sequence | Or | Xor | And |
          Concurrent | Not | Any | Aperiodic |  Periodic
```

Event Algebra Operators

```
operator ::= Sequence | Or | Xor | And | Concurrent | Not | Any | Aperiodic |
          Periodic | Cterm
Sequence ::= [oid,] {event | Ind | Var | Cterm | operator | Sequence | Or | Xor
             | And | Concurrent | Not | Any | Aperiodic |  Periodic }
Or ::= [oid,] {event | Ind | Var | Cterm | operator | Sequence | Or | Xor
       | And | Concurrent | Not | Any | Aperiodic |  Periodic }
And ::= [oid,] {event | Ind | Var | Cterm | operator | Sequence | Or | Xor
        | And | Concurrent | Not | Any | Aperiodic |  Periodic }
Xor ::= [oid,] {event | Ind | Var | Cterm | operator | Sequence | Or | Xor
        | And | Concurrent | Not | Any | Aperiodic |  Periodic }
Concurrent::= [oid,] {event | Ind | Var | Cterm | operator | Sequence | Or |
             Xor  | And | Concurrent | Not | Any | Aperiodic |  Periodic }
Not::= [oid,] event | Ind | Var | Cterm | operator | Sequence | Or | Xor  | And |
   Concurrent | Not | Any | Aperiodic |  Periodic , interval | Interval | Plex |
Var
Any ::= [oid,] Ind | Data | Var , event | Ind | Var | Cterm | operator | Se-
quence
       | Or | Xor  | And | Concurrent | Not | Any | Aperiodic |  Periodic
Aperiodic ::= [oid,] event | Ind | Var | Cterm | operator | Sequence | Or | Xor
|
           And | Concurrent | Not | Any | Aperiodic |  Periodic, interval |
          Interval | Plex | Var
Periodic ::= [oid,] time | Ind | Var | Cterm, interval | Interval | Plex | Var
```

Although the EC axioms might be represented directly in plain RuleML syntax, we decided to provide additional more specialzed constructs on the EC layer to represent the axioms and event algebra operators directly and in a more compact and natural way. For example an EC fact happens(e,t) can be serialized in RuleML as an Atom, e.g.

```
<Atom>
          <Rel>happens</Rel>
          <Ind>e</Ind>
          <Ind>t</Ind>
</Atom>
```

However, a more compact and intuitive representation is reached with the new EC constructs which reveal their semantics directly:

```
<Happens>
        <event><Ind>e</Ind></event>
        <time><Ind>t</Ind></time>
</Happens
```

To represent the event algebra operators directly in RuleML we might use the reified style of nested complex terms to model and combine the algebra operators. This approach using Cterms will enable us to define complex event detection patterns directly in an ECA rule instead of implementing them in derivation rules which are queried by the ECA rules' event part.

Examples

RuleML

Sequence operator (;): (a;b):

```
<Cterm><Ctor>sequence</Ctor><Ind>b</Ind><Ind>b</Ind></Cterm>
```

Negation operator (¬): (¬b [a,c]):

```
<Cterm><Ctor>not</Ctor><Ind>b</Ind><Cterm><Ctor>interval</Ctor><Ind
>a</Ind><Ind>c</Ind></Cterm></Cterm>
```

Quantification (Any): (Any(n,a);c):

```
<Cterm><Ctor>sequence</Ctor><Cterm><Ctor>any</Ctor><Data>3</Data
><Ind>a</Ind></Cterm><Ind>c</Ind></Cterm>
```

ECA-RuleML

Sequence operator (;): (a;b):

```
<Sequence><Ind>b</Ind><Ind>b</Ind></Sequence>
```

Negation operator (¬): (¬b [a,c]):

```
<Not><Ind>b</Ind><Interval><Ind>a</Ind><Ind>c</Ind></Interval></Not>
```

Quantification (Any): (Any(n,a);c):

```
<Sequence><Any><Ind>3</Ind><Ind>a</Ind></Any><Ind>c</Ind></Sequence>
```

ECA-RuleML is not intended to be executed directly, but transformed into the target execution language of an underlying rule-based systems (e.g. a Prolog interpreter or a Rete based forward reasoning engine) and then executed there. We have implemented XSLT stylesheets which transform RuleML and ECA-RuleML into ContractLog LPs [6] and Prova scripts [7]. ContractLog is an expressive KR framework developed in the RBSLA project[8] which is based on Prova / Mandarax, an open-source Java-based rule engine with a Prolog like scripting syntax. The ContractLog KR implements several logical formalisms such as event logics, defeasible logic, deontic logics and the formalisms and meta programs described in section 3 and 4 have been implemented within this KR. Similar mappings via XSLT to other languages and event algebras such as SNOOP are possible.

[6] http://ibis.in.tum.de/staff/paschke/rbsla/contractlog.htm

[7] http://www.prova.ws/

[8] http://ibis.in.tum.de/staff/paschke/rbsla/index.htm

# 6. Performance Evaluation

In this section we will evaluate our approach by means of asymptotic worst case complexity analysis and experimental performance tests as well as proof-of-concept examples derived from common industry use cases.

## 6.1. Theoretical Worst-Case Complexity

We first introduce the necessary terminology:
A relation r(t) on the set of ground terms of a LP language is definable under $\models$ semantics if there exists a program P and predicate symbol p(t) in the language such that for every ground t, r(t) ≡ P $\models$ r(t) or r(t) ≡ P $\models$ ¬r(t). We further distinguish a database of facts $D_{in}$ and a set of input rules P for inferring additional information. More precisely the predicate symbols in a LP language are divided into a set of extensional relations and a set of intensional relations. The facts $D_{in}$ are formed from the extensional predicates and ground terms, while the rule heads are formed from the intensional predicates. The worst case complexity is given by the complexity of checking whether $P \cup D_{in} \models A$ for a variable logic program P, ground atoms A, and input database $D_{in}$. Fixing the intensional relations leads to data complexity which is the complexity of checking whether a ground atom A is entailed by $P \cup D_{in}$ for a fixed set of intensional predicates and a variable set of extensional predicates, i.e. a fixed program P and a variable input factbase $D_{in}$. Data complexity can be viewed as a function of the size of $D_{in}$. The program complexity is the complexity of checking $P \cup D_{in} \models A$ for variable programs P and ground atoms A over a fixed input database $D_{in}$.

**ECA Processor / ECA Rules**

Algorithm 6.1: Operational semantics of ECA processor

*1.* The ECA processor collects all ECA rules (represented as facts) from the KB via a query *eca(T,E,C,A,P,EL)?*
2. Then it processes each constituent term in each ECA rule according to the forward directed ECA paradigm.

Theorem 6.1: *The forward directed execution of ECA rules within the ECA processor takes linear time O(n) where n is the number of ECA rules.*

Proof 6.1: (sketch)
Step (1) is P-complete under logspace reductions:
All ECA rules are represented as ground facts in the KB. Hence, the complexity $D_{in} \cup P \models A$ is equivalent to the data complexity of datalog for a variable input database $D_{in}$ and a fixed datalog program P, where the set of ECA rules $E := \{eca_1,..,eca_n\} \subseteq D_{in}$. Datalog has been proven to be data complete in P, in contrast to program complexity which is DEXPTIME, e.g. implicit in [67]. The problem of query answering on (untyped) ground facts (without rules) reduces to the problem of unifying the query with the facts which is P-complete under logspace reductions (see e.g. [68]) and O(n) in the ContractLog KR for all facts in E. The output $D_{out}$ has size O(n) since ECA rules are represented as facts in the KB and the query is completely unbound, i.e. all terms are variables which leads to a full instantiation of terms via unification.

Note: The hybrid typed logic (Description Logic or Java type system) [69] with a typed unification used in ContractLog extends the decision problem of deciding whether two typed terms are unifiable to dynamic sub-type checking under the type order with ad-hoc polymorphism for variables. This is done by dynamic class checks in case of Java class hierarchies and by description logics (DL) inferences (subsumption and instantiation) in case of RDFS / OWL type vocabularies (ontologies). Accordingly, for hybrid DL-typed LPs with Datalog restriction (is typically fulfilled for ECA rules) the worst-case complexity is dependent on the DL language *L* used to represent the type system, i.e. complexity of EXPTIME for OWL-Lite reps. NEXPTIME for OWL-DL. As a result, although, typed unification using DL types is decidable, it has rather high worst-case complexity. Nevertheless, there are highly optimized DL reasoners such as Fact, Racer, Pellet available, which behave adequately in practice.

Step (2) takes O(n) time since operationally selecting an empty thread and processing the constituent parts takes the same amount of time for each rule. Note the time to evaluate the ECA parts is neglected here since the evaluation time of the ECA rules' parts is dependent on their underlying implementation, e.g. procedural (via procedural attachment), LP-based (query on derivation rules) and differs significantly according to the LP semantics (e.g. well founded, stable model, answer set) and LP class (e.g. definite Datalog LP, normal LP, extended LP, disjunctive LP) of the underlying rule system and rule language.

**Classical Event Calculus**

Algorithm 6.2: Query answering in the basic Event Calculus (holdsAt)

1. First all occurred events have to queried*: happens(Ei,T1)?*
2. Then all events from *Ei* which happened before the queried time *T* need to be selected: *T1<t*
3. Then all events *Ei* which initiate the queried fluent need to be selected: *initiates(Ei,F,T1)*?
4. And finally it needs to be checked whether there exists another event *Et* which terminates the fluent between *T1* and *T*: *notclipped(T1,Fluent,T)?*.

Theorem 6.2: *In the case of absolute times and total ordering of events the worst case data complexity of query answering whether a fluent holds or not at a given time point has linear complexity* O(n) *where n is the number of events.*

Proof 6.2: (sketch)

The general complexity of the EC is the complexity of checking whether a certain fluent holds *holdsAt(Fluent,t)?* based on the happened events and the defined EC inference rules initiating respectively terminating this fluent. Hence, the cost is measured by data complexity for a fixed set of initiates/terminates rules and a variable set of happens facts $D_{in}$ as the number of accesses to the knowledge base to unify the happens facts during the refutation attempt. The complexity is given as a function of the number n of occurred events.

Step (1) takes linear time O(n) and has O(n) output size. The query unifies directly with the happens facts and succeeds n times as much events are recorded in the KB. Hence, the complexity reduces to the unification problem and has cost O(n) (see proof 6.1).

Since we impose absolute times and total orderings of events step (2) is a simple date comparison function (solved via procedural code) which has constant cost.

Step (3) takes O(n), since the initiates resp. terminates are typically represented as facts which a queried by a bound query (i.e. reduces to the unification problem – see proof 6.1).

Step (4) again needs to query all events *n* which cost *O(n)* to derive *Et* and compare the occurrence dates and the termination rules with constant costs.

According to this the overall complexity is also linear *O(n)* by the number of events *n*.

**Interval-based Event Calculus**

The interval-based Event Calculus, i.e. query answering with holdsInterval on the set of occurred volatile events, has similar complexity than the basic event calculus since an event interval is spanned by an initiating and a terminating event which both have to be looked up in the database and then other terminating events which possibly break the interval must be searched, which is comparable to the clipped auxiliary rule in the basic event calculus. Since the central time concept in the interval-based EC are time intervals and not single time points the comparison of the time intervals is a little bit more expensive than comparing two single time points, but the overall asymptotic cost is still linear.

## 6.2. Experimental Evaluation

We ran the performance tests on an Intel Pentium 1,2 GHz PC with 512 MB RAM running Windows XP. The first test evaluates the ECA processor, the second the Event Calculus. We use as a measure of problem size the total number of ECA rules (test 1) and the total number of occurred event facts (test 2) in the KB.

Scalable ECA Test Theories

The basic ECA test theories $eca_{basic}(n)$ consists of *n* ECA rules *eca(_,_,_,_)*, where the ECA parts have no functionality, but lead to a full execution of the ECA rule. The rules are processed sequentially by the ECA processor. A variant $eca_{basic^*}(n)$ uses multi-threading and executes the rule in parallel. The ECA daemon first queries the KB for new ECA rules via a query "*eca(T,E,C,A,P,El)?*" and populates the active KB with them. Then it evaluates the ECA rule one after another. Accordingly, update time for querying the KB for new/updated ECA rules and processing time for executing all ECA rules must be distinguished

$$eca_{basic}(n) = \begin{cases} eca_1: eca(_,_,_,_,_,_). \\ \ldots \\ eca_n: eca(_,_,_,_,_,_). \end{cases}$$

Scalable Event Calculus Test Theories

In the basic EC test theories $ec_{basic}(n)$ there is a pair of EC rules initiating / terminating a property (fluent) $p$ by means of $n$ alternating events (happens).

$$ec_{basic}(n) = holdsAt(p_m,t_n) = \begin{cases} initiates(e_1,p,T) \\ terminates(e_2,p,T) \\ happens(e_1,t_0) \\ happens(e_2,t_1) \\ \ldots \\ happens(e_1,t_n) \end{cases}$$

Table 1 shows the experiment results in CPU milliseconds.

| | Problem Size | Time | |
|---|---|---|---|
| | | Update | Execution |
| $eca_{basic}(n)$ | 1000 | 0.7 | 0.005 |
| | 2500 | 1.6 | 0.01 |
| | 5000 | 3.1 | 0.02 |
| | 10.000 | 6.3 | 0.035 |
| $ec_{basic}(n)$ | 22 | 0.18 | |
| | 42 | 0.37 | |
| | 82 | 0.71 | |
| | 162 | 1.37 | |

The experiments reveal adequate performance results, e.g. the EC handles an event instance sequence (EIS) with 82 "unconsumed" event facts used to detect a complex event in less than 0.7 seconds, i.e. decides whether the fluent $p$ initiated by $e1$ and terminated by $e2$ with 41 happened $e1$ facts and 41 happened $e2$ facts holds. Due to our analysis in the SLA domain (see Paschke, A., Schnappinger-Gerull, E.: A Categorization Scheme for SLA Metrics**.** Multi-Conference Information Systems (MKWI06), Passau, Germany, 2006) typical event instance sequences for one complex event detection are much smaller with less than ten contributing events.

### 6.3. Industrial Use Case

In this section we evaluate our approach towards event/action processing and reactive rules with respect to epistemological adequacy by means of a formalization of a typical use case found in industry in ECA-LP which is one possible, target execution syntax for the ECA-RuleML markup language:

**Example:** ContractLog ECA-LP formalization of a SLA

"*The service availability will be measured every $t_{schedule}$ according to the actual schedule by a ping on the service. If the service is unavailable and it is not maintenance then escalation level 1 is triggered and the process manager is informed. Between 0-4 a.m. the process manager is permitted to start servicing which terminates any escalation level. The process manager is obliged to restart the service within time-to-repair, if the service is unavailable. If the process manager fails to restore the service in time-to-repair (violation of obligation), escalation level 2 is triggered and the chief quality manager is informed. The chief quality manager is permitted to extend the time-to-repair interval up to a defined maximum value in order to enable the process manager to restart the service within this new time-to-repair. If the process manager fails to restart the service within a maximum time-to-repair escalation level 3 is triggered and the control committee is informed. In escalation level 3 the service consumer is permitted to cancel the contract.*"

The formalization in ContractLog ECA-LP is as follows:

```
% service definition
service(http://ibis.in.tum.de/staff/paschke/rbsla/index.htm).
% role model and escalation levels
initially(escl_lvl(0)).  % initially escalation level 0
role(process_manager) :- holdsAt(escl_lvl(1),T). % if escalation level 1 
then process_manager
role(chief_quality_manager) :-  holdsAt(escl_lvl(2),T). % if escalation 
level 2 then chief quality manager
role(control_committee) :- holdsAt(escl_lvl(3),T). % if escalation level 3 
then control committee
% time schedules standard, prime, maintenance and monitoring 
intervals
% before 8 and after 18 every minute
schedule(standard, Service):-
        systime(datetime(Y,M,D,H,Min,S)), 
less(datetime(Y,M,D,H,Min,S), datetime(Y,M,D,8,0,0)),
        interval(timespan(0,0,1,0), datetime(Y,M,D,H,Min,S)),
        service(Service), not(maintenance(Service)).  % not mainte-
nance
schedule(standard, Service):-
        systime(datetime(Y,M,D,H,Min,S)), 
more(datetime(Y,M,D,H,Min,S), datetime(Y,M,D,18,0,0)),
        interval(timespan(0,0,1,0), datetime(Y,M,D,H,Min,S)),
        service(Service), not(maintenance(Service)).  % not mainte-
nance
% between 8 and 18 every 10 seconds
schedule(prime, Service):-
        sysTime(datetime(Y,M,D,H,Min,S)),
        lessequ(datetime(Y,M,D,H,Min,S),datetime(Y,M,D,18,0,0)),
        moreequ(datetime(Y,M,D,H,Min,S),datetime(Y,M,D,8,0,0)),
         interval(timespan(0,0,0,10), datetime(Y,M,D,H,Min,S)) , ser-
        vice(Service).
% between 0 and 4 if maintenance every 10 minutes
schedule(maintenance, Service) :-
        sysTime(datetime(Y,M,D,H,Min,S)), lesse-
        qu(datetime(Y,M,D,H,Min,S),datetime(Y,M,D,4,0,0)),
        interval(timespan(0,0,10,0), datetime(Y,M,D,H,Min,S)) ,
        service(Service), maintenance(Service). % servicing
initiates(startServicing(S),maintenance(S),T).        % initiate mainte-
nance if permitted
```

terminates(stopServicing(S), maintenance(S),T). % terminate maintenance
happens(startServicing(Service),T):-
happens(requestServicing(Role,Service),T), holdsAt(permit(Role,Service, startServicing(Service)),T).
% ECA rule: "If the ping on the service fails and not maintenance then trigger escalation level 1 and notify process manager, else if ping succeeds and service is down then update with restart information and inform responsible role about restart".
eca(schedule(T,S), not(available(S)), not(maintenance(S)), escalate(S),_, restart(S)). % ECA rule

available(S) :- WebService.ping(S). % ping service
maintenance(S) :- sysTime(T), holdsAt(maintenance(S),T).
escalate(S) :- sysTime(T), not(holdsAt(unavailable(S),T)), % escalate only once
add("outages","happens(outage(_0),_1).",[S,T]),% add event
role(R), notify (R, unavailable(S)). % notify
restart(S) :- sysTime(T), holdsAt(unavailable(S),T),
add("outages","happens(restart(_0),_1).",[S,T]),% add event
role(R), notify(R,restart(S)). % update + notify
% initiate unavailable state if outage event happens
initiates(outage(S),unavailable(S),T). terminates(restart(S),unavailable(S),T).
% initiate escalation level 1 if outage event happens
terminates(outage(S),escl_lvl(0),T). initiates(outage(S),escl_lvl(1),T).
% terminate escalation level 1/2/3 if restart event happens
initiates(restart(S),escl_lvl(0),T). terminates(restart(S),escl_lvl(1),T).
terminates(restart(S),escl_lvl(2),T). terminates(restart(S), escl_lvl(3), T).
% terminate escalation level 1/2/3 if servicing is started
initiates(startServicing(S),escl_lvl(0),T). terminates(startServicing(S), escl_lvl(1),T). terminates(startServicing(S), escl_lvl(2),T). terminates(startServicing(S),escl_lvl(3),T).
% permit process manager to start servicing between 0-4 a.m.
holdsAt(permit(process_manager,Service, startServicing(Service)), datetime(Y,M,D,H,Min,S)):-
lessequ(datetime(Y,M,D,H,Min,S),datetime(Y,M,D,4,0,0)).
% else forbid process manager to start servicing.
holdsAt(forbid(process_manager,Service, startServicing(Service)), datetime(Y,M,D,H,Min,S)):-
more(datetime(Y,M,D,H,Min,S),datetime(Y,M,D,4,0,0))..
% derive obligation to start the service if service unavailable
derived(oblige(process_manager, Service , restart(Service))). % oblige process manager
holdsAt(oblige(process_manager, Service , restart(Service)), T) :- holdsAt(unavailable(Service),T).
% define time-to-repair deadline and trigger escalation level 2 if deadline is elapsed
time_to_repair($t_{deadline}$). % relative time to repair value
trajectory(escl_lvl(1),T1,deadline,T2,(T2 - T1)) . % deadline function
derivedEvent(elapsed).
happens(elapsed,T) :- time_to_repair(TTR), valueAt(deadline,T, TTR).
terminates(elapsed, escl_lvl(1),T).% terminate escalation level 1
initiates(elapsed, escl_lvl(2),T). % initiate escalation level 2
% trigger escalation level 3 if (updated) time-to-repair is > max time-to-repair
happens(exceeded,T) :- happens(elapsed,T1), T=T1+ $ttr_{max}$.
terminates(exceeded,escl_lvl(2),T). initiates(exceeded, escl_lvl(3),T).
% service consumer is permitted to cancel the contract in escl_lvl3
derived(permit(service_consumer, contract , cancel)).
holdsAt(permit(service_consumer, contract , cancel), T) :- holdsAt(escl_lvl(3),T).

Via simple queries the actual escalation level and the rights and obligations each role has in a particular state can be derived from the rule base and the maximum validity interval (MVI) for each contract state which is managed by the EC formalization, e.g. the maximum outage time, can be computed. These MVIs can be used to compute the service levels such as average availability. The ECA processor of the ContractLog framework actively processes the ECA rules. Every $t_{check}$ according to the actual schedule it pings the service via a procedural external call, triggers the next escalation level if the service is unavailable and informs the corresponding role. As can be seen from this use case example, the declarative rule based approach allows a very compact representation of ECA rules in combination of globally defined derivation rules, which would not be possible in the standard procedural approaches in the active database domain.

## 7. Related Work

Related work can be divided into work on active event processing and event algebras in the active database community and work on event/action logics, updates, state processing/transitions and temporal reasoning in the knowledge representation domain. In the former several prototypes have been developed, e.g. ACCOOD [5], Chimera [6], ADL [7], COMPOSE [8], NAOS [9], HiPac [10], AMIT [11], RuleCore[9]. These systems mainly treat complex event detection purely procedural or on a simple level and focus on specific aspects. Other papers discuss formal aspects on a more general level - see [14] for an overview.

There has been also a lot of research and development concerning knowledge updates and active rules (execution models) in the area of active databases and several techniques based on syntactic (e.g. triggering graphs [17] or activation graphs [18]) and semantics analysis (e.g. [19], [20]) of rules have been proposed to ensure termination of active rules (no cycles between rules) and confluence of update programs (always one unique outcome). The combination of deductive and active rules has been also investigated in different approaches manly based on the simulation of active rules by means of deductive rules [21-23, 70]. However, in contrast to our work these approaches often assume a very simplified operational model for active rules without complex events and ECA related event processing.

Several event algebras have been developed, e.g. Snoop [4], SAMOS [16], ODE [8]: The object database ODE [8] im-

[9] ruleCore: http://www.rulecore.com

plements event-detection mechanism using finite state automata. SAMOS [16] combines active and object-oriented features in a single framework using colored Petri nets. Associated with primitive event types are a number of parameter-value pairs by which events of that kind are detected. SAMOS does not allow simultaneous atomic events. Snoop [4] is an event specification language which defines different restriction policies that can be applied to the operators of the algebra. Complex events are strictly ordered and cannot occur simultaneously. The detection mechanism is based on trees corresponding to the event expressions, where primitive event occurrences are inserted at the leaves and propagated upwards in the trees as they cause more complex events to occur. A feature that is common to all these event algebras is that event operators are the only way to specify the semantics of complex events and as we have shown in this paper due to the treatment of complex events as occurrences at a single point in time, rather than over an interval, some operators show several irregularities and inconsistencies.

In the area of knowledge representation events have been also studied in different areas. Several Datalog extensions such as the LDL language of Naqvi and Krishnamurthy [56] which extends Datalog with update operators including an operational semantics for bulk updates or the family of update language of Abiteboul and Vinau [57] which has a Datalog-style have been proposed. A number of further works on adding a notion of state to Datalog programs where updates modelled as state transitions close to situation calculus has been taken place [22, 55]. The situation calculus [62] is a methodology for specifying the effects of elementary actions in first-order logic. Reiter further extended this calculus with an induction axiom specified in second-order logic for reasoning about action sequences. Applying, this Reiter has developed a logical theory of database evolution [71], where a database state is identified with a sequence of actions. This might be comparable to the Event Calculus, however, unlike the EC the situation calculus has no notion of time which in many event-driven systems is crucial, e.g. to define timestamps, deadline principles, temporal constraints, or time-based context definitions. Transaction Logics [31] is a general logic of state changes that accounts for database updates and transactions supporting order of update operations, transaction abort and rollback, savepoints and dynamic constraints. It provides a logical language for programming transactions, for updating database views and for specifying active rules, also including means for procedural knowledge and object-oriented method execution with side effects. It is an extension of classical predicate calculus and comes with its own proof theory and model theory. The proof procedure executes logic programs and updates to databases and generates query answers, all as a result of proving theorems. Although it has a rich expressiveness in particular for combining elementary actions into complex ones it primarily deals with database updates and transactions but not with general event processing functionalities in style of ECA rules such as situation detection, context testing in terms of state awareness or external inputs/outputs in terms of event/action notifications or external calls with side effects.

Multi-dimensional dynamic logic programming [58], LUPS [72], EPI [60], Kabul [61] Evolp [59], ECA-LP [35] are all update languages which enable intensional updates to rules and define a declarative and operational semantics of sequences of dynamic LPs. While LUPS and EPI support updates to derivation rules, Evolp and Kabul are concerned with updates of reaction rules. However, to the best of our knowledge only ECA-LP [35] supports a tight homogenous combination of reaction rules and derivation rules enabling transactional updates to both rule types with post-conditional (ECAP) verification, validation, integrity tests (V&V&I) on the evolved knowledge state after the update action(s). Moreover, none of the cited update languages except of ECA-LP supports complex (update) actions and events as well as external events/actions. Moreover these update languages are only concerned on the declarative evolution of a knowledge base with updates, whereas the Event Calculus as used in ECA-RuleML/ECA-LP is developed to reason about the effects of actions and events on properties of the knowledge systems, which hold over an interval.

Concerning mark-up languages for reactive rules several proposals for update languages exist, e.g. XPathLog [73], XUpdate [74], XML-RL [75] or an extension to XQuery proposed by Tatarinov et.al. [76]. This has been further extended with general event-driven and active functionalities for processing and reasoning about arbitrary events occurring in the Web and other distributed systems as well as triggering actions. Most of the proposals for an ECA or reactive web language are intended to operate on local XML or RDF databases and are basically simply trigger approaches, e.g. Active XML [77], Active Rules for XML [78], Active XQuery [79] or the ECA language for XML proposed by Bailey et.al [80] as well as RDFTL for RDF [81]. XChange [82] is a high level language for programming reactive behaviour and distributed applications on the Web. It allows propagation of changes on the Web (change) and event-based communication (reactivity) between Web-sites (exchange). It is a pattern-based language closely related to the web query language Xcerpt with an operational semantics for ECA rules of the form: "*Event query* → *Web query* → *Action*". Events are represented as XML instances that can be communicated and queried via XChange event messages. Composite events are supported, using event queries based on the Xcerpt query language [83] extended by operators similar to an event algebra applied in a tree-based approach for event detection.

## 8. Conclusion

In this paper we have discussed two major approaches in event and action processing, namely reactive rules from ac-tive databases and event/action logics from the KR domain. We have compared their differences in view of complex event and action definitions, which either treat complex events in terms of instantaneous occurrences as in the active database community or as durative interval-based events or event sequences as in the event notification and KR domain. We have shown that the former style has several shortcomings regarding unintended semantics of some of their event algebra operators and a restricted expressiveness as compared to declarative programming approaches such as logic programming. We have presented an approach to unifying these approaches using the framework of logic programming as a common representation basis for refining typical event algebra operators in terms of interval based complex events. In particular, we have developed an event algebra based on a KR event logics formalisms, the interval-based Event Calculus, and we have shown how the forward directed operational semantics of ECA rules can be effectively implemented within a backward-reasoning LP environment and combine with our EC event/action algebra. The main benefits of this combination are:

- Based on a precise, declarative logic-based semantics (see ECA-LP [35])
- Highly expressive and computationally feasible (linear worst-case complexity O(n) in the untyped framework)
- Reactive rules might be represented and used in combination with other rule types, such as derivation rules (e.g. business rules) or integrity constraints, within the same framework of logic programming
- Might be easily combined and extended with other logical formalisms (e.g. defeasible/default for conflict handling) to build reliable high-level decision logics upon.
- The inference engine is used as generic interpreter to detect complex events and enable immediate but also delayed and inferred reactions and effects
- The ECA interpreter works as an add-on to arbitrary backward-reasoning engines which can be easily integrated by a Wrapper interface on the query API of the underlying rule engine

We have proposed a light-weight and compact ECA Rule Markup Language (ECA-RuleML). The ECA-RuleML reuses the rich syntax of derivation rules already defined in RuleML. Hence it fits perfectly into the language family of RuleML and allows easy combination of reactive rules and derivation rules, e.g., combination of business rules with event-driven active rules. It fulfills typical criteria of good language design, such as orthogonality, lean set of constructs, and homogeneity. We have demonstrated adequacy of this approach by means of theoretical and experimental evaluation and common examples derived from real-world industry use cases. The approach is in particular well suited for applications where the ability to reason formally about the system is crucial for describing reliable and verifiable real-world decision logic and triggering reactions in response to occurred complex events and event sequences.

## Further Information / Links

The implementations of the active LP-based ECA processor and the logical formalisms and meta programs for the basic event calculus, interval-based event calculus, interval-based event algebra, (transactional) ID-based update primitives, integrity constrains, test cases described in section 3, are part of the *ContractLog KR*. The ContractLog KR is an expressive and efficient KR framework hosted at Sourceforge for the representation of contractual rules implementing several logical formalisms such as event logics, defeasible logic, deontic logics, description logic programs etc. The ContractLog KR is developed within the *Rule Based Service Level Agreement (RBSLA) project.*

*ECA-RuleML* is a sublanguage of the *Rule Based Service Level Agreement language (RBSLA)* which comprises further layers adding further modelling power such as deontic norms or defeasible rules with rule priorities in order to adequately represent Service Level Agreements:

RBSLA project: http://ibis.in.tum.de/research/rbsla/index.htm
RBSLA distribution: https://sourceforge.net/projects/rbsla
ContractLog: http://ibis.in.tum.de/research/rbsla/contractlog.htm
ECA-RuleML: http://ibis.in.tum.de/research/eca-ruleml
RBSLA language: http://ibis.in.tum.de/research/rbsla/rbsla.htm

Further information on RBSLA and ContractLog:

- Paschke, A., RBSLA: A Rule Based Service Level Agreements Language based on RuleML. IBIS, TUM, Technical Report 8/2004.
- Paschke, A.: ContractLog - A Logic Framework for SLA Representation, Management and Enforcement. IBIS, TUM, Technical Report, 07/2004.
- Paschke, A.: RBSLA - *A declarative Rule-based Service Level Agreement Language based on RuleML*, International Conference on Intelligent Agents, Web Technology and Internet Commerce (IAWTIC 2005), Vienna, Austria, 2005.
- Paschke, A.: *Typed Hybrid Description Logic Programs with Order-Sorted Semantic Web Type Systems based on OWL and RDFS*, Internet Based Information Systems, Technical University Munich, Technical Report 12/05, 2005.
- Paschke, A.: *ECA-LP: A Homogeneous Event-Condition-Action Logic Programming Language*, IBIS, Technische Universität München, Technical Report 11 / 2005.

The ContractLog KR and ECA-LP is implemented on top of Prova as a reference rule engine.. Prova is an open source Java-based backward-reasoning rule engine derived from Mandarax. It provides a Prolog related scripting sysntax and combines Java (via Java reflection) and derivation rules:

Prova: http://www.prova.ws/

# Appendix A: ECA RuleML Syntax (Part)

Procedural Attachment
Cterm ::= [oid,] op | Ctor | Attachment, {slot,} [resl,] {arg | Ind | Data | Skolem | Var | Reify | Cterm | Plex}, [repo], {slot}, [resl]
Attachment ::= [oid,] Ind | Var | Cterm, Ind

ID based update constructs
Assert ::= content | And
Retract ::= content | And
RetractAll ::= content | And

ECA Rule
ECA ::= [oid,] [time,] [event,] [condition,] action [,postcondition] [,else]
time ::= Naf | Neg | Cterm
event ::= Naf | Neg | Cterm | Sequence | Or | Xor | And | Concurrent | Not | Any | Aperiodic | Periodic
condition ::= Naf | Neg | Cterm
action ::= Cterm | Assert | Retract | RetractAll
postcondition ::= Naf | Neg | Cterm
else ::= Cterm | Assert | Retract | RetractAll

Naf ::= [oid,] weak | Atom | Cterm
Neg ::= [oid,] strong | Atom | Equal | Cterm

Event Calculus
event ::= Ind | Var | Cterm
fluent ::= Ind | Var |Cterm
parameter ::= Ind | Var |Cterm
time :: = Ind | Var | Cterm
interval::=Interval | Plex | Var

Happens ::= [oid,] event | Ind | Var | Cterm , time | Ind | Var | Cterm
Planned ::= [oid,] event | Ind | Var | Cterm, time | Ind | Var | Cterm
Occurs ::= [oid,], event | Ind | Var | Cterm, interval | Interval | Plex | Var
Initially ::= [oid,] fluent | Ind | Var | Cterm
Initiates ::= [oid,] event | Ind | Var | Cterm, fluent | Ind | Var | Cterm, time | Ind | Var | Cterm
Terminates ::= [oid,] event | Ind | Var | Cterm, fluent | Ind | Var | Cterm | interval | Interval, time | Ind | Var | Cterm | interval | Interval
HoldsAt ::= [oid,] fluent | Ind | Var | Cterm , time | Ind | Var | Cterm
ValueAt ::= [oid,] parameter | Ind | Var | Cterm, time | Ind | Var | Cterm, Ind | Var | Cterm | Data
HoldsInterval ::= [oid,] interval | Interval | Plex | operator | Sequence | Or | Xor | And | Concurrent | Not | Any | Aperiodic | Periodic | Cterm , interval | Interval | Plex | Var
Interval::= [oid,] event | time | Ind | Var | Cterm | operator | Sequence | Or | Xor | And | Concurrent | Not | Any | Aperiodic | Periodic , event| time | Ind | Var | Cterm | operator | Sequence | Or | Xor | And | Concurrent | Not | Any | Aperiodic | Periodic

Event Algebra Operators
operator ::= Sequence | Or | Xor | And | Concurrent | Not | Any | Aperiodic | Periodic | Cterm
Sequence ::= [oid,] {event | Ind | Var | Cterm | operator | Sequence | Or | Xor | And | Concurrent | Not | Any | Aperiodic | Periodic }
Or ::= [oid,] {event | Ind | Var | Cterm | operator | Sequence | Or | Xor | And | Concurrent | Not | Any | Aperiodic | Periodic }
And ::= [oid,] {event | Ind | Var | Cterm | operator | Sequence | Or | Xor | And | Concurrent | Not | Any | Aperiodic | Periodic }
Xor ::= [oid,] {event | Ind | Var | Cterm | operator | Sequence | Or | Xor | And | Concurrent | Not | Any | Aperiodic | Periodic }
Concurrent::= [oid,] {event | Ind | Var | Cterm | operator | Sequence | Or | Xor | And | Concurrent | Not | Any | Aperiodic | Periodic }
Not::= [oid,] event | Ind | Var | Cterm | operator | Sequence | Or | Xor | And | Concurrent | Not | Any | Aperiodic | Periodic , interval | Interval | Plex | Var
Any ::= [oid,] Ind | Data | Var , event | Ind | Var | Cterm | operator | Sequence | Or | Xor | And | Concurrent | Not | Any | Aperiodic | Periodic
Aperiodic ::= [oid,] event | Ind | Var | Cterm | operator | Sequence | Or | Xor | And | Concurrent | Not | Any | Aperiodic | Periodic, interval | Interval | Plex | Var
Periodic ::= [oid,] time | Ind | Var | Cterm, interval | Interval | Plex | Var

# Appendix B – ECA-RuleML Example

ECA
eca(schedule(T,S), not(available(S)), not(maintenance(S)), escalate(S),_, restart(S)). % ECA rule

is serialized to

```
<ECA>
 <time>
      <Cterm>
             <Ctor>schedule</Ctor>
             <Var>T</Var>
            <Var>S</Var>
       </Cterm>
 </time>
  <event>
     <Naf>
      <Cterm>
              <Ctor>available</Ctor>
              <Var>S </Var>
      </Cterm>
    </Naf>
  </event>
  <condition>
     <Naf>
        <Cterm>
           <Ctor>maintenance</Ctor>
          <Var>S</Var>
        </Cterm>
     </Naf>
  </condition>
 <action>
         <Cterm>
                 <Ctor>escalate</Ctor>
                 <Var>S</Var>
       </Cterm>
  </action>
  <else>
        <Cterm>
                 <Ctor></Ctor>
                 <Var>S</Var>
        </Cterm>
  </else>
</ECA>
```

<u>Event Calculus</u>

initiates(startServicing(S),maintenance(S),T).
terminates(stopServicing(S), maintenance(S),T).

is serialized to

```
<Initiates>
  <event>
     <Cterm>
              <Ctor>startServicing</Ctor>
              <Var>S</Var>
      </Cterm>
  <event>
  <fluent>
      <Cterm>
              <Ctor>maintenance</Ctor>
              <Var>S</Var>
       </Cterm
  </fluent>
  <time>
       <Var>T</Var>
  <time>
</Initiates>
<Terminates>
      <Cterm>
              <Ctor>stopServicing</Ctor>
             <Var>S</Var>
       </Cterm>
      <Cterm>
              <Ctor>maintenance</Ctor>
              <Var>S</Var>
        </Cterm
        <Var>T</Var>
</Terminates>
```

## Appendix C – Overview RuleML 0.9

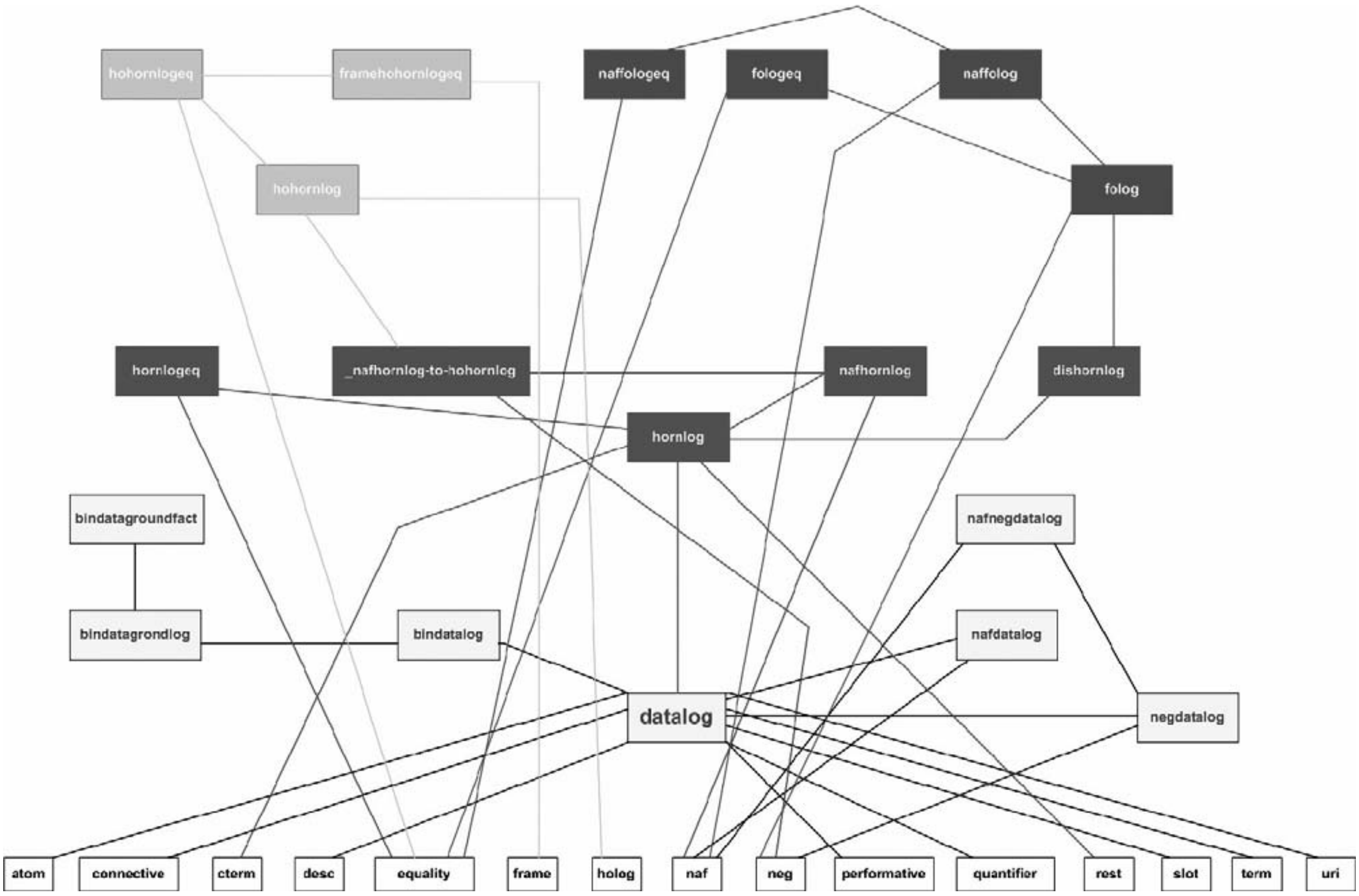

## Appendix D: Basic Event Calculus (Part) - holdsAt

```
% Optimized version with cut to answer bound queries holdsAt(f,t)?
holdsAt(Fluent,T):-
        bound(Fluent),
        bound(T), % T must be input / bound
        initiates(AnEvent, Fluent, T),
        happens(AnEvent, Before),
        less(Before, T),
        notclipped(Before, Fluent, T),  % assumes closed world
        !.

% Optimized version to answer free queries holdsAt(F,t)?
holdsAt(Fluent,T):-
        bound(T), % T must be input / bound
        happens(AnEvent, Before),
        less(Before, T),
        initiates(AnEvent, Fluent, Before),
        notclipped(Before, Fluent, T).  % assumes closed world

% Auxiliary predicate
clipped(T1, Fluent, T2) :-
        terminates(AnEvent, Fluent, T2),
        happens(AnEvent, AnTime),
        lessequ(T1, AnTime),
        less(AnTime, T2).
```

## Appendix E: Interval-based Event Calculus Meta Program (Part)

```
% -----------------------------------------------------------------------
% Atomic event occurences and complex event occurences
% with occurence interval [T1,T2]
% -----------------------------------------------------------------------
event([Event],T):- occurs(Event,T). % atomic event
% complex event defined by event algebra operators
event([Operator|Events],[T1,T2]):-
        operator(Operator),
        derive(Operator(Events,[T1,T12],T2,Events)).
event([[Operator|Events]],[T1,T2]):-
        operator(Operator),
        derive(Operator(Events,[T1,T12],T2,Events)).
```

```
%-----------------------------------------------------------------------------------------------
% holdsInterval - Optimized version for queries with free time intervals
% Returns the all time intervals [T1,T2] where the event interval [E1,E2]
% occurs and is not terminated in between. The terminators are defined
% explicitly, i.e.  globally be a terminates predicate. For local terminators
use
% the holdsInterval variant with local terminators.
% E1 and E2 must be bound, i.e. E1 initiates the interval and E2 terminates
% it. Hence no EC initates predicate is needed.
%-------------------------------------------------------------------------------------------------
holdsInterval([E1,E2],[T11,T22]):-
        free(T11), free(T22),
        event([E1],[T11,T12]),
        event([E2],[T21,T22]),
        lessequ([T11,T12], [T21,T22]),
        notbroken(T12, [E1,E2], T21).

%-------------------------------
% Interval with one event.
%----------------------------------
holdsInterval([E],[T1,T2]):-

        event([E],[T1,T2]).

% ---------------------------------------
% broken(T1, Interval, T2)
% tests if the interval [E1,E2] is terminated between T1 and T2 by an
% terminating event
% ---------------------------------------
broken(T1, Interval, T2) :-
        terminates(AnEvent, Interval, T2),
        event([AnEvent],[T11,T12]),
        less(T1, T11),
        less(T12, T2),

        !.

% -------------------------------------------------------------------------
% Sequence Operator with local terminators derived from event sequence
% -------------------------------------------------------------------------
sequence([E],[T1,T2],T2,Terminators):-
        holdsInterval([E],[T1,T2]).
sequence([E1|Rest],[T1,T2],Tend, Terminators):-
        head(E2,Rest),
        holdsInterval([E1,E2],[T1,T2],Terminators),
        sequence(Rest,[T2,T3],Tend,Terminators),
        lessequ([T1,T2],[T2,T3]).
```

## Appendix F: Integrity Constraints - testIntegrity(Literal)

```
% ----------------------------------------------------------------
% ContractLog defines four types of integrity constraints
% and:   intgertiy(and(...)).
% or:   intgertiy(or(...)).
% xor:   intgertiy(xor(...)). (mutual exclusive)
% not:   intgertiy(not(...)).
% ----------------------------------------------------------------
% test Not integrity constraints
testIntegrity(Literal):-
        integrity([not|NotConstraints]),
        member(Literal,NotConstraints),!,fail().

% test XOR (mutex) integrity constraints
testIntegrity(Literal):-
        integrity([xor|Mutex]),
        delete(Literal,Mutex,NMutex),
        member([H|T],NMutex),
        derive([H|T]),!,fail().

% test And integrity constraints
testIntegrity(Literal):-
        integrity([and|AndConstraints]),
        delete(Literal,AndConstraints,NAndConstraints),
        member(M,NAndConstraints),
        deriveLiteral(M),!,fail().
```